\RequirePackage{fix-cm}
\documentclass[twocolumn]{svjour3}          
\smartqed  
\usepackage{xcolor}
\usepackage{tabularx}
\usepackage{multirow}
\usepackage{graphicx}
\usepackage{makecell}
\usepackage{amsmath}
\usepackage{color}
\usepackage{float}
\usepackage{rotating}
\usepackage{placeins}
\usepackage{amsfonts}
\usepackage{bm}
\usepackage[title]{appendix}

\usepackage[round,authoryear,comma,sort&compress,numbers]{natbib} 
\usepackage[colorlinks=true,linkcolor=blue,citecolor=black,urlcolor=blue]{hyperref}

\usepackage{diagbox}
\usepackage{cancel} 
\usepackage{mathrsfs} 
\usepackage{eqnarray} 

\usepackage{tikz}
\usepackage{ctable} 

\usepackage{algorithm}
\usepackage{algpseudocode}%
\usepackage{listings}%
\usepackage{bbding}
\usepackage{bbm}
\usepackage[caption=false, font=footnotesize]{subfig}
\usepackage{fontawesome}
\usepackage{array}

\usepackage{colortbl}  
\usepackage{arydshln}
\usepackage{longtable}
\usepackage{wrapfig}
\usepackage{bbold}
\usepackage{pifont}

\usepackage{xcolor}
\usepackage{microtype}
\hypersetup{hidelinks,
	colorlinks=true,
	allcolors=black,
	pdfstartview=Fit,
	breaklinks=true}

\definecolor{lime}{HTML}{A6CE39}
\DeclareRobustCommand{\orcidicon}{
\begin{tikzpicture}
\draw[lime, fill=lime] (0,0)
circle[radius=0.13]
node[white]{{\fontfamily{qag}\selectfont \tiny \.{I}D}};
\end{tikzpicture}
\hspace{-2mm}
}
\foreach \x in {A, ..., Z}{%
\expandafter\xdef\csname orcid\x\endcsname{\noexpand\href{https://orcid.org/\csname orcidauthor\x\endcsname}{\noexpand\orcidicon}}
}

\newcommand{\PreserveBackslash}[1]{\let\temp=\\#1\let\\=\temp}
\newcolumntype{C}[1]{>{\PreserveBackslash\centering}p{#1}}
\newcolumntype{R}[1]{>{\PreserveBackslash\raggedleft}p{#1}}
\newcolumntype{L}[1]{>{\PreserveBackslash\raggedright}p{#1}}

\definecolor{battleshipgrey}{rgb}{0.52, 0.52, 0.51}
\definecolor{capri}{rgb}{0.0, 0.75, 1.0}
\definecolor{mediumspringgreen}{rgb}{0.0, 0.98, 0.6}
\definecolor{Gray}{rgb}{0.7,0.7,0.7}

\def\ourmodel{Brain3D}

\journalname{%
  \parbox[t]{8cm}{%
    International Journal of Computer Vision
  }%
}
\begin{document}

\title{Brain3D: Generating 3D Objects from fMRI}

\author{
	Yuankun Yang$^1$ \and
	Li Zhang$^1$\hspace{-2mm}\orcidA{}\hspace{-2mm} \and
    Ziyang Xie$^2$ \and
    Zhiyuan Yuan$^1$ \and
	Jianfeng Feng$^1$ \and
    Xiatian Zhu$^3$ \and
       Yu-Gang Jiang$^1$ 
}

\institute{
	Corresponding author: Li Zhang  \at
             \email{lizhangfd@fudan.edu.cn}          \\
$^1$ Fudan University \\
$^2$ University of California, Los Angeles \\
$^3$ University of Surrey
}
\date{5th Mar 2025}

\maketitle

\begin{abstract}
Understanding the hidden mechanisms behind human's visual perception is a fundamental question in neuroscience.
To that end, investigating into the neural responses of human mind activities,
such as functional Magnetic Resonance Imaging (fMRI),
has been a significant research vehicle.
However, analyzing fMRI signals is challenging, costly,
daunting, and demanding for professional training.
Despite remarkable progress in fMRI analysis, existing approaches are limited to generating 2D images and far away from being biologically meaningful and practically useful. 
Under this insight, 
we propose to generate visually plausible and functionally more comprehensive 3D outputs decoded from brain signals,
enabling more sophisticated modeling of fMRI data.
Conceptually, we reformulate this task as
a {\em fMRI conditioned 3D object generation} problem.
We design a novel 3D object representation learning method, \ourmodel, that takes as input the fMRI data of a subject who was presented with a 2D image,
and yields as output the corresponding 3D object images.
The key capabilities of this model include tackling the noises
with high-level semantic signals and a two-stage architecture design for progressive high-level information integration.
Extensive experiments validate the superior capability of our model over previous state-of-the-art 3D object generation methods.
Importantly, we show that our model captures the distinct functionalities of each region of human vision system
as well as their intricate interplay relationships, aligning remarkably with the established discoveries in neuroscience.
Further, preliminary evaluations indicate that \ourmodel{} can successfully identify the disordered brain regions in simulated scenarios, such as V1, V2, V3, V4, and the medial temporal lobe (MTL) within the human visual system. Our data and code will be available at 
\href{https://brain-3d.github.io/}{https://brain-3d.github.io/}.
\end{abstract}

\begin{figure*}[!ht]
    \centering
\includegraphics[width=\linewidth]{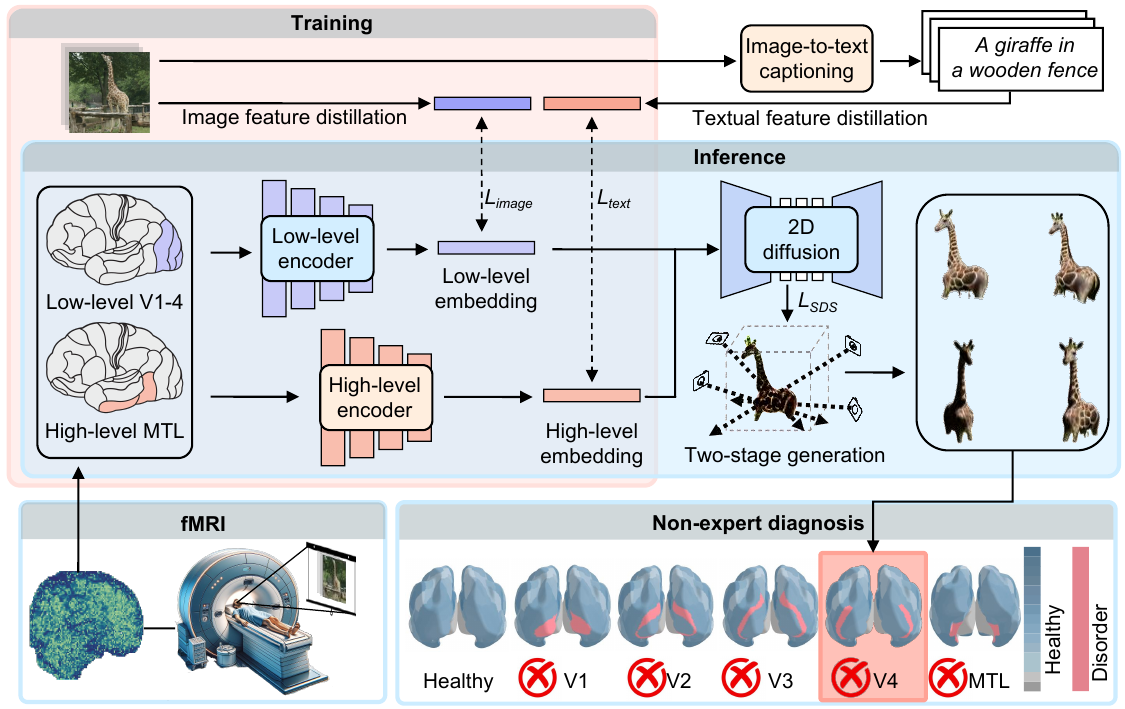}
\caption{\textbf{\ourmodel: Generating 3D objects from brain fMRI signals.} 
    A subject's visual system first precepts object stimulus, triggering specific patterns of activity in the brain's visual processing areas which is then captured and expressed by fMRI data. \ourmodel{} then leverages a fMRI encoder to extract the feature, feeds it into
    a 2D diffusion model for detailed 3D generation. 
    The training procedure involves optimizing low-level and high-level encoders with the feature distillation objective $L_{image}$ and $L_{text}$ respectively, each incorporating bidirectional CLIP~\citep{Radford2021clip} contrastive learning and MixUp~\citep{Zhang2017mixup}  augmentation for enhanced generalization.
    By contrasting the object stimulus and generated 3D visuals, our approach offers the potential for exploring brain region functionalities and aiding in the diagnosis of neurological conditions.
    }   
    \label{fig:1}
\end{figure*}

\section{Introduction}
Understanding how humans see and perceive the world is a fundamental question in computer vision~\cite{Hubel1979brain, Hupé1998cortial}. The ability to perceive and interpret the environment in three dimensions (3D) is a critical aspect of human vision that enables human to navigate, interact, and make sense of the world around us~\cite{Georgieva2009,Groen2019}. Despite significant advancements in the understanding of the neural mechanisms underlying visual perception~\cite{Tong2003primary, Popham2021}, the intrinsic processes involved in 3D vision remain largely unexplored.

Traditional approaches in neuroscience are limited by ethical concerns~\cite{Díaz2021}, interspecies differences~\cite{Jansen2020}, and a lack of high-resolution temporal and spatial data, which hinders the translation of findings to human vision. Similarly, while artificial vision models have excelled in computer vision tasks~\cite{Kubilius2019,Sexton2022,Baek2021,Frey2023}, they fall short in mimicking the neural processes of biological vision and fail to incorporate brain signals effectively. Neural decoding models have made strides in reconstructing visual content from fMRI data~\cite{Seeliger2018,Shen2019,Chen2023, Scotti2024}, yet they are constrained by their focus on single views, limiting their ability to fully capture the complexity of human 3D vision.

Consequently, our goal here is to develop a comprehensive computing framework that bridges the gap between artificial vision models and biological vision and gains insights into human visual system, with a specific focus on understanding and simulating the human brain's 3D visual processing capabilities. 
To that end, here we explore the intrinsic 3D human vision perception through the lens of fMRI data~\cite{allen2022massive}, known for its high spatial resolution, relatively low cost compared to other neuroimaging techniques, and non-invasiveness. We introduce Brain3D, a bio-inspired method capable of understanding how the brain interprets and processes visual stimuli in 3D. Brain3D takes fMRI data of a subject presented with a 2D object image and yields the corresponding 3D object model. 
There are two key components with Brain3D: tackling the noises
with high-level semantic signals by decomposing it into a series of representative embedding vectors, and a two-stage architecture design for progressive high-level information integration.
Importantly, our model captures the distinct functionalities and interplay relationships of each region in the human vision system, aligning with established neuroscience discoveries. This alignment allows Brain3D to potentially serve as a useful tool for clinical fMRI evaluation.

Our {\bf contributions} are summarized as follows:

\textbf{(i)}
We introduce Brain3D, a bio-inspired method that translates fMRI into 3D objects, advancing the understanding of human 3D vision perception. This research pioneers the understanding of the neural basis of human 3D vision with artificial neural networks, shedding light on cognitive processes and paving the way for theoretical exploration and practical applications in neuroscience.

\textbf{(ii)}
We develop specialized neural encoders with a two-stage generation process to accurately model the distinct functionalities and interplay of regions in the human visual system, building on established neuroscience findings.

\textbf{(iii)}
We demonstrate Brain3D's ability to investigate regional functionalities, including the left and right brain hemispheres, various visual regions, and the medial temporal lobe (MTL), highlighting its potential for practical applications in clinical fMRI evaluation.
Preliminary experiments demonstrate that our method can effectively identify regions associated with disorders, including V1, V2, V3, V4, and MTL, offering the opportunity for innovative diagnostic and therapeutic approaches in neurology and psychology.

\section{Related work}

\noindent\textbf{Neuroscience}
Traditional approaches in neuroscience and life sciences, such as animal studies~\cite{Costa2022,Chen2022} and invasive techniques~\cite{Singh2022,Risom2022}, have provided valuable insights into neural mechanisms. However, these methods face significant limitations, including the need for animal sacrifice~\cite{Rai2018}, ethical concerns~\cite{Díaz2021}, and interspecies heterogeneity~\cite{Jansen2020}, which can hinder the translation of findings to human vision. Additionally, traditional methods often lack the spatial and temporal resolution necessary to capture the complex dynamics of neural activity underlying 3D vision perception. Our work offers a novel interdisciplinary approaches that explore intrinsic human vision perception through non-invasive fMRI-based generation.

\noindent\textbf{Computer vision}
Computer vision,
which aims to develop computational models inspired by biological vision~\cite{KhalighRazavi2014, Lecun2015DeepLearning}, has emerged as a promising avenue for understanding the neural basis of visual perception. While vision models have achieved remarkable success in various tasks, such as 
object recognition~\cite{Kubilius2019,Sexton2022}, detection~\cite{Baek2021}, and scene understanding~\cite{Frey2023}, their contribution to understanding biological vision remains limited. These models primarily focus on catching human-like performance on specific tasks using visual input, without directly incorporating brain signals or mimicking the neural processes involved in biological vision. Furthermore, these models are not designed to simulate the human visual system.
In contrast, this work directly integrates brain signals into a visual computing model, enabling the simulation of 3D vision that aims to reflect the neural mechanisms of biological vision.

\noindent\textbf{Neural decoding}
Neural decoding models have focused on extracting essential representations of brain signals for visual content prediction~\cite{Naselaris2011, Kamitani2005} and object recognition~\cite{Horikawa2017}.
These investigations have also shown promise in reconstructing images~\cite{Beliy2019} from fMRI data using techniques such as generative adversarial networks~\cite{Seeliger2018,Shen2019} and latent-space diffusion~\cite{Chen2023, Scotti2024}. 
However, these studies are currently limited to single views, which constrains their ability to model and analyze the underlying human brain's 3D visual processing capabilities.  When considering brain-signal-based 3D synthesis, the intrinsically significant diversity within the human brain can cause instability, resulting in a lack of consistency for 3D synthesis.
To bridge this gap, our work pioneers the use of fMRI data for generating 3D object models, capturing the complex interplay between different regions of the human visual system.

\noindent\textbf{Cross-modality 3D generation}
Previous advancements in cross-modality 3D generation are limited in
text-based~\citep{poole2022dreamfusion, lin2023magic3d, wang2024prolificdreamer, tang2023dreamgaussian} and image-based~\citep{liu2023zero, shi2023mvdream, qianmagic123} pipelines. These generation approaches leverage score distillation sampling~\citep{poole2022dreamfusion} (SDS) and extensive 3D datasets~\citep{deitke2023objaverse, yu2023mvimgnet} to distill text or image into 3D representations, including NeRF~\citep{Ben2020NeRF}, DMTet~\citep{Shen2021dmtet} or Gaussian splatting~\citep{kerbl20233d}. In comparison, our work builds on the concept of hybrid SDS~\cite{qianmagic123} to distill brain signals into both high-level semantics and low-level visuals. This distinctive capability allows our model to achieve cross-modal 3D generation from solely brain signals, even without external images or texts as conditional guidance.

\section{Method}

\begin{figure*}[!t]
    \centering
\includegraphics[width=\linewidth]{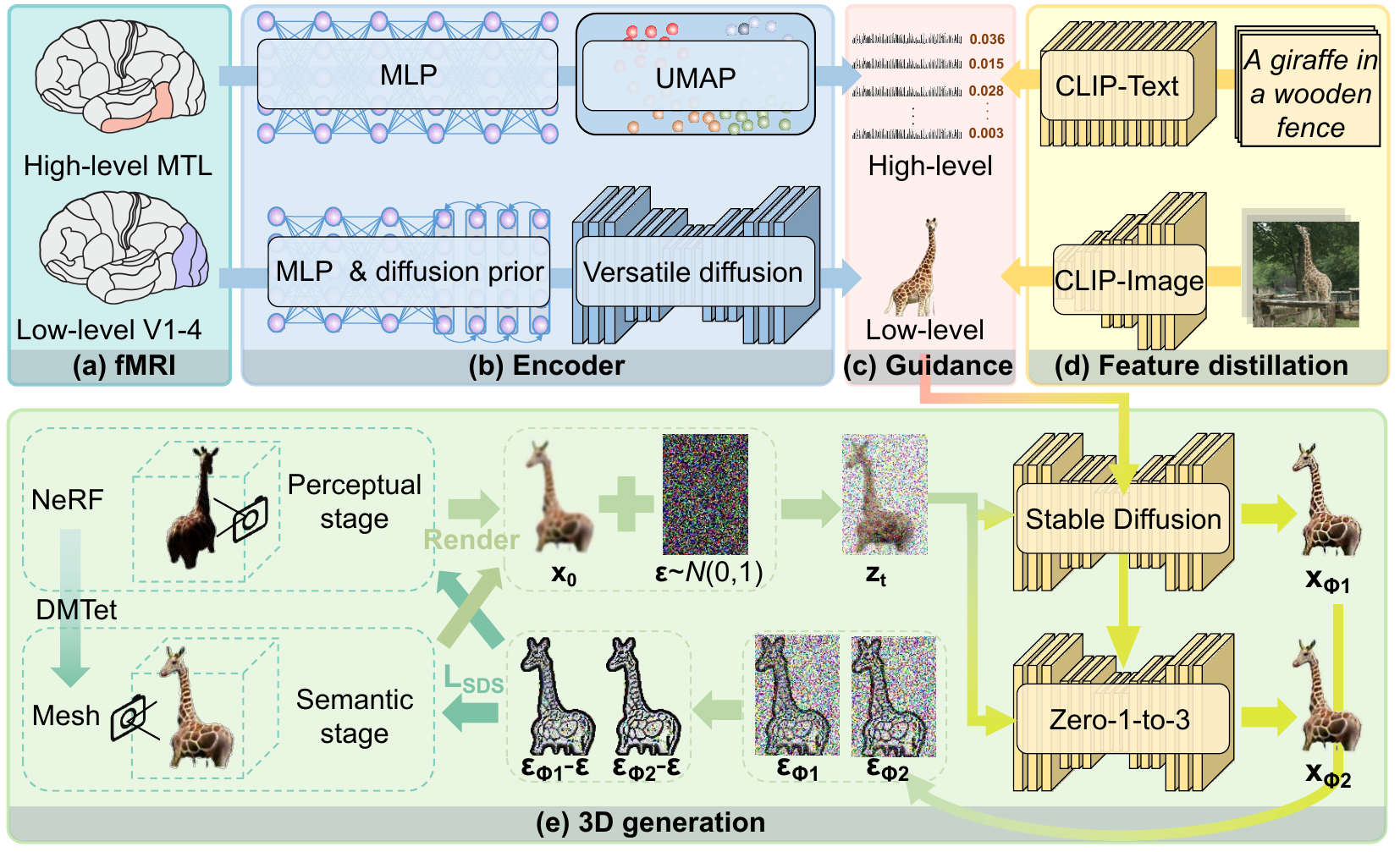}
    \caption{
\textbf{Overview of our \ourmodel{} that decodes the functional MRI (fMRI) signals into 3D reconstructions.} (a) Initially, the subject's visual system precepts the object stimulus, which triggers specific patterns of neural activity within the brain's visual processing areas. These patterns are recorded as fMRI signals.  (b) To generate the 3D object, \ourmodel{} employs two specialized encoders: the high-level encoder uses MLP and UMAP~\cite{mcinnes2018umap} to capture abstract, high-level visual concepts, while the low-level encoder processes basic visual details via Versatile diffusion~\cite{Xu2023}. (c) Both encoders process fMRI to extract respective information for representing 3D objects. (d) These encoders are optimized through feature distillation, which utilize pretrained CLIP-Text and CLIP-Image encoders~\cite{Radford2021clip} as the guidance for the high-level and low-level features, respectively. (e) The 3D object generation process includes two phases: the perceptual phase utilizes NeRF~\cite{Ben2020NeRF}, and the semantic phase transforms NeRF output into a 3D mesh~\cite{Hoppe1993mesh} using DMTet~\cite{Shen2021dmtet}. The 3D object is refined using Score Distillation Sampling (SDS)~\cite{poole2022dreamfusion}, which introduces Gaussian noise into a randomly-viewed rendered image and employs 2D diffusion models for noise prediction. The high-level and low-level feature guidance both are served as conditions for the pretrained Stable Diffusion~\cite{rombach2022high} and Zero-1-to-3~\cite{liu2023zero-1-to-3} models to predict the added noise.
}
    \label{fig:2}
\end{figure*}

\subsection{Overview}
\label{sec:method_overview}

As a bio-inspired method our \ourmodel{} can understand how the brain interprets and processes visual stimuli in 3D 
 (Fig.~\ref{fig:2}). 
It efficiently maps 2D visual stimuli into 3D objects by analyzing the intricate functionalities of distinct brain regions. 
Specifically, our model first deploys two specialized neural encoders to encode fMRI data into multi-level neural embeddings(Fig.~\ref{fig:2}a-d). 
The low-level encoder is designed to encode fMRI data from the basic visual cortex (V1 to V4), focusing on visual details such as color, shape, and texture, into low-level neural embeddings. Concurrently, the high-level encoder concentrates on the medial temporal lobe (MTL) to process visual information, such as object semantics and types. It encodes fMRI data from the MTL regions to high-level embeddings.

Our key technical contribution lies in the novel design of a UMAP-based projection~\cite{mcinnes2018umap}, which effectively addresses high-level semantic noise, allowing us to decompose these embeddings into separate distinct representations. This enhancement is crucial for deriving 3D multi-view semantics from brain signals, thereby facilitating accurate 3D generation.
Subsequently, we design a two-stage diffusion method~\cite{ho2020denoising,rombach2022high,saharia2022photorealistic} to reconstruct the corresponding 3D objects based soley on brain signals. 
3D objects are generated through distilling a 2D diffusion model with condition on both the high level and low level fMRI embeddings, using Score Distillation Sampling (SDS)~\cite{poole2022dreamfusion}(Fig.~\ref{fig:2}e). 
The results affirm \ourmodel{}'s effectiveness in modeling human vision process and diagnosing neurological abnormalities across multiple brain regions, including V1, V2, V3, V4, and MTL. 

\subsection{Preliminary: Diffusion based 3D generation} 
\label{sec:sup_Preliminary}
Our approach employs a diffusion model~\cite{ho2020denoising} aimed at manipulating data through a two-phased process. Initially, a forward process introduces Gaussian noise with parameter $(\alpha_t, \sigma_t)$ into the data $\mathbf{x}_0$ to achieve noised input $\mathbf{z}_t$, which can be mathematically represented as follows:
\begin{equation}
\resizebox{\dimexpr\linewidth-25pt\relax}{!}{
$
q_t(\mathbf{z}_t|\mathbf{x}_0)=N(\alpha_t \mathbf{x}_0,\sigma_t^2 I)=\int q_t(\mathbf{z}_t|\mathbf{x}_0) q_0(\mathbf{x}_0)d\mathbf{x}_0.
$
\label{eq:forward diffusion}}
\end{equation}
Following this, a reverse process is employed to denoise the data, starting from pure noise. This reverse mechanism is driven by a parameterized noise prediction network $\epsilon_{\phi}(\mathbf{x}_t,t)$ , optimized by minimizing the objective loss function as:
\begin{equation}
\resizebox{\dimexpr\linewidth-25pt\relax}{!}{
$L_{\rm{Diff}}(\phi,\mathbf{x}) = \mathbb{E}_{t,\epsilon } \left[ \omega(t) \left| \epsilon_\phi(\alpha_t\mathbf{x}+\sigma_t \epsilon, t)-\epsilon_t \right|_2^2 \right].$
}
\label{eq:backward diffusion}
\end{equation}
This mechanism is further extended to 3D generation through the integration of SDS~\cite{poole2022dreamfusion}. It involves two key components: a scene model and a pre-trained Text-to-Image (T2I) diffusion model. 
The scene model, parameterized by $\theta$, generates differentiable images through $\mathbf{X_i} = g(\theta)$. The pre-trained T2I diffusion model predicts the sampled noise $\epsilon_{\phi}(\mathbf{X_i}_t; t, \mathbf{y})$ conditioned on the noisy image $\mathbf{X_i}_t$, created by adding Gaussian noise with a random scale factor $t$ to $\mathbf{X_i}$. The optimization of this extended model is guided by the following objective:
\begin{equation}
\resizebox{\dimexpr\linewidth-25pt\relax}{!}{$
\begin{aligned}
&\nabla_{\theta}\mathcal{L}_{SDS}(\phi, \mathbf{X_i}) = \mathbb{E}_{t,\epsilon} \left[\omega(t)
\left(\hat{\epsilon}_{\phi}(\mathbf{X_i}_t; t, \mathbf{y}) - \epsilon_t\right)
\frac{\partial \mathbf{X_i}}{\partial \theta}\right] \\
&= \mathbb{E}_{t,\mathbf{z}_t | \mathbf{X_i}} \left[ \omega(t) \frac{\sigma_t}{\alpha_t} \nabla_{\theta} \mathrm{KL} \left(q(\mathbf{z}_t | \mathbf{X_i} = g(\theta)) \, || \, p_{\phi}(\mathbf{z}_t | \mathbf{y})\right) \right].
\end{aligned}
$}
\label{eq:partial_SDS}
\end{equation}
Through this computation, our model seeks to align the generated distribution  $q(\mathbf{z}_t |\mathbf{X_i}=g(\theta))$ closely with the target distribution $ p_{\phi}(\mathbf{z}_t |\mathbf{y})$, ensuring consistency across various camera poses, as detailed in the following integration:
\begin{equation} p_{\phi}(\mathbf{z}_t |\mathbf{y})=\int p_{\phi}(\mathbf{z}_t |c,\mathbf{y})p(c) dc.
\label{eq:integration}
\end{equation}
This integration necessitates consistency in $p_{\phi}(\mathbf{z}_t |c,\mathbf{y})$ from each camera pose $c$.

\subsection{Tackling the noises with high-level semantics} 
\label{sec:UMAP}

\begin{figure*}[!t]
    \centering
        \parbox{0.1\linewidth}{\centering  \footnotesize (a)Image } 
        \parbox{0.28\linewidth}{\centering  \footnotesize (b)Embedding w.o UMAP }        
        \parbox{0.30\linewidth}{\centering  \footnotesize (c)UMAP projected embedding (i) }         
        \parbox{0.30\linewidth}{\centering  \footnotesize (d)UMAP projected embedding (ii)} 

    \includegraphics[width=\linewidth]{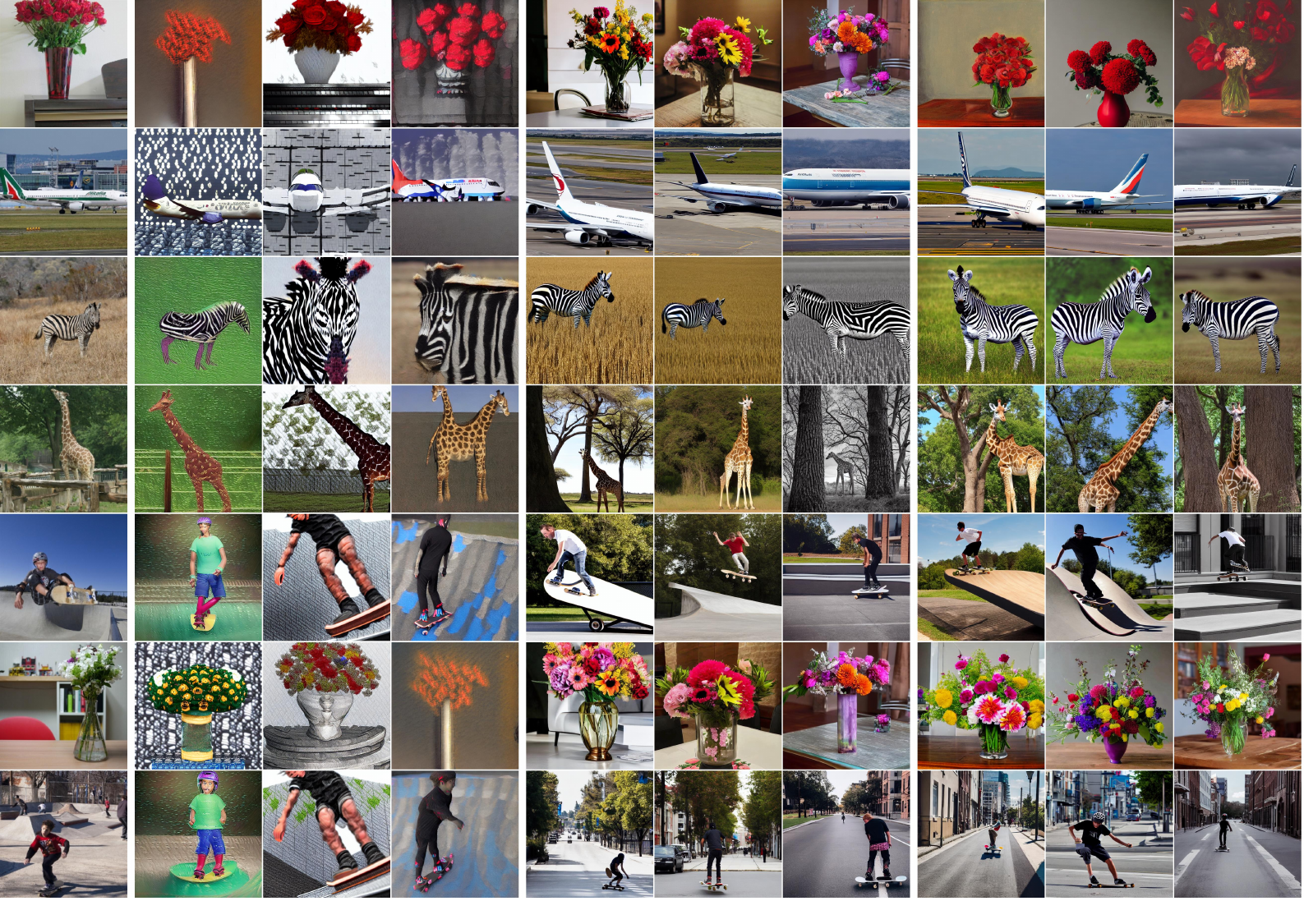}
    \caption{\textbf{Variety in high-level information.} (a) The reference images presented to the participants. 
    (b) Images generated by Stable Diffusion\cite{rombach2022high} conditioned on the high-level embedding without UMAP\cite{mcinnes2018umap}  projection.
    (c, d) Images conditioned on different high-level embeddings after UMAP projection.
    It shows that the high-level embedding directly extracted from fMRI comes with high diversity and noise. Our UMAP projection can significantly mitigate this challenge.
}
    \label{high level}
\end{figure*}

In modeling human mind signals, a major challenge arises from the intrinsically significant levels of noise and diversity with the high-level information within the human brain (Fig.~\ref{high level}). 
This can cause instability in the conditional probability distribution $p_{\phi}(\mathbf{z}_t |c,\mathbf{y})$ (see Sec.~\ref{sec:sup_Preliminary} for more details), particularly when considering different camera angles, denoted as $c$. This instability could result in a lack of consistency for 3D object generation from the complex fMRI scans.

To solve the above problem, we design a novel approach using the Uniform Manifold Approximation and Projection (UMAP)~\cite{mcinnes2018umap} (Fig.~\ref{fig:2}b).
Unlike MindEye using UMAP solely for visualization, we incorporate UMAP projection into our model architecture to effectively manage high-level semantic noise.
The key idea is to distill the high-level semantic embeddings into a series of representative embedding vectors with a unique projection.

Formally, 
following the transformation of fMRI $\mathbf{X}_i$ into a high-level embedding by the MLP backbone $f_H$, this embedding $f_H(\mathbf{X}_i)$ is subsequently projected 
into a set of stable embeddings,
$\left\{ f_{H_1}(\mathbf{X}_i), f_{H_2}(\mathbf{X}_i), \ldots, f_{H_n}(\mathbf{X}_i) \right\}$, derived from the Generative Text-to-Image Transformer~\cite{wang2022git}.
In our design, we concatenate the embeddings extracted from all training images. This results in a comprehensive set of embeddings, $\left\{ f_{H}(\mathbf{X}_i), f_{H_1}(\mathbf{X}_i), f_{H_2}(\mathbf{X}_i), \ldots, f_{H_n}(\mathbf{X}_i) \right\}$, for each image.  These combined embeddings form a high-dimensional dataset, denoted as $\textbf{D}$.

Then, we optimize the stable representation of these embeddings, with the goal to minimize the cross-entropy between the high-dimensional probability distributions and their corresponding stable embeddings. It is conducted via stochastic gradient descent for handling the complexity of such high-dimensional data.

For fMRI inference $\mathbf{X}_i$, we deploy a pre-trained UMAP model, denoted as $g_U$, to project the high-level embeddings $\left\{ f_{H}(\mathbf{X}_i), f_{H_1}(\mathbf{X}_i), f_{H_2}(\mathbf{X}_i), \ldots, f_{H_n}(\mathbf{X}_i) \right\}$ into a stable space. The resultant embeddings are represented as $\left\{\mathbf{U}_H, \mathbf{U}_{H_1}, \mathbf{U}_{H_2},..., \mathbf{U}_{H_n}\right\}$ with each obtained as:
\begin{equation}
\mathbf{U}_{H_j}=g_U(f_{H_j}(\mathbf{X}_i)).
\label{eq:projection}
\end{equation} 
 To quantify the similarity between $\mathbf{U}_H$ and each $\mathbf{U}_{H_j}$, we calculate the cosine similarity, $\rm{s_i}$, as follows:
\begin{equation}
\rm{s_i} = \frac{{\mathbf{U}_H \cdot \mathbf{U}_{H_j}}}{{|\mathbf{U}_H| \cdot |\mathbf{U}_{H_j}|}},
\label{eq:cosine}
\end{equation} 
This similarity measurement is then normalized to establish a sampling probability for each embedding:
\begin{equation}
\rm{v}_j = \frac{\rm{s}_j}{| \mathbf{U}_{H_j} |},
\label{eq:normalize}
\end{equation} 
\begin{equation}
\rm{w}_j=\frac{\rm{v}_j}{\sum_{j=1}^n \rm{v}_j}, 
\label{eq:normalize2}
\end{equation} 
This normalization process is critical as it determines how embeddings are sampled during the denoising step of the 2D diffusion model. Under specific conditions, the linear space formed by $\left\{\mathbf{U}_{H_1}, \mathbf{U}_{H_2},..., \mathbf{U}_{H_n}\right\}$ can effectively represent $\mathbf{U}_H$ linearly, as described in the follows:
\begin{equation}
\resizebox{\dimexpr\linewidth-25pt\relax}{!}{$
\begin{aligned}
\mathbf{U}_H &= \sum_{j=1}^n \mathrm{proj}_{\mathbf{U}_{H_j}}(\mathbf{U}_H) = \sum_{j=1}^n \frac{\mathbf{U}_{H_j} \cdot \mathbf{U}_{H}}{| \mathbf{U}_{H_j} |^2}  \mathbf{U}_{H_j} \\
&= \sum_{j=1}^n s_j \cdot \frac{|\mathbf{U}_{H}|}{\mathbf{U}_{H_j}} \mathbf{U}_{H_j} = |\mathbf{U}_{H}| \sum_{j=1}^n t_j \cdot  \mathbf{U}_{H_j}.
\end{aligned}$}
\label{Eq:UMAP}
\end{equation}
In essence, the UMAP projection serves as a non-linear expression of the high-level embedding $f_H(\mathbf{X}_i)$, contributing significantly to our model's ability to handle complex data representations.

In diffusion model's inference step, we sample the projected embedding $f_{H_j}(\mathbf{X}_i)$ with probability $w_j$ for denoising. This approach ensures a stable $p_{\phi}(\mathbf{z}_t |\mathbf{y}_i)=\int p_{\phi}(\mathbf{z}_t |c,\mathbf{y}_i)p(c) dc$. Consequently, the gradients of the loss function $\mathcal{L}_{SDS}(\phi, \mathbf{X}_i)$ are modified as follows:
\begin{equation}
\resizebox{\dimexpr\linewidth\relax}{!}{$
\begin{aligned}
&\nabla_{\theta}\mathcal{L}_{SDS}(\phi, \mathbf{X}_i)
=\mathbb E_{t,i,\epsilon} [\omega(t)(\hat{\epsilon}_{\phi}(\mathbf{X}_{i_t};t,\mathbf{y_i})-\epsilon_t) \frac{\partial \mathbf{X}_i}{\partial \theta}] \\
&=E_{t,i,\mathbf{z}_t | \mathbf{X}_i} [ \omega(t) \frac{ \sigma_t}{\alpha_t} \nabla_{\theta} KL (q(\mathbf{z}_t |\mathbf{X}_i=g(\theta)  || p_{\phi}(\mathbf{z}_t |\mathbf{y}_i))) ],
\label{eq:partial_SDS_Modified}
\end{aligned}$}
\end{equation}
Here, the expectation of the sampled embedding $i$ is integrated into the computation.
This means we selectively utilize these weights to pick the most relevant high-level embeddings.
This integration is crucial for ensuring that our model accurately captures and reconstructs the complex patterns inherent in fMRI data, producing more reliable and consistent multi-view visual results.

\subsection{Two-stage 3D generation}
To effectively leverage the projected high-level embeddings, we devise a two-stage 3D generation architecture characterized by progressively integrating high-level information during 3D object modeling(Fig.~\ref{fig:2}e). The first stage is a perceptual stage, utilizing the Neural Radiance Fields (NeRF)~\cite{Ben2020NeRF} for 3D representation. It defines an implicit 3D representation framework as:
\begin{equation} 
C(\textbf{r})=\int_{t_n}^{t_f} T(t)\sigma(\textbf{r}(t))\textbf{c}(\textbf{r}(t),\textbf{d})dt, 
\label{eq:Nerf}
\end{equation}
\begin{equation} 
T(t)=\rm{exp}(-\int_{t_n}^{t}\sigma(\textbf{r}(s))ds).
\label{eq:Nerf2}
\end{equation}
This framework allows the gradients from the Score Distillation Sampling (SDS)~\cite{poole2022dreamfusion} to directly influence the output $C(\mathbf{r})$, enabling simultaneous optimization of the density Multilayer Perceptron (MLP) $\sigma(\mathbf{r}(t))$ and color MLP $(\mathbf{r}(t),\mathbf{d})$. As a result, NeRF is particularly effective at modifying both the color and shape of the 3D object in this stage. In our implementation, we incorporate Zero-1-to-3~\cite{liu2023zero-1-to-3} for low-level guidance and Stable Diffusion~\cite{rombach2022high} for high-level guidance.

Subsequently, a semantic stage follows where we employ DMTet~\cite{Shen2021dmtet} to convert the above NeRF into a 3D mesh~\cite{Hoppe1993mesh}. This conversion prepares the 3D model for further processing.  We continue to harness Stable Diffusion~\cite{rombach2022high}, coupled with a depth-conditioned Control-Net~\cite{zhang2023adding} for high-level guidance. While the 3D mesh does not possess the same versatility as NeRF in shaping modifications, it excels in capturing detailed RGB color information. Therefore, after establishing the coarse shape in the perceptual stage, we shift our focus to refining the color details in this stage. This two-tier design allows for an intricate and nuanced development of the 3D object, leading to a final output that is both elaborately detailed and of high resolution.

\subsection{Training and inference}

During training, we optimize the parameters for both low-level encoder and high-level encoder, leaving all the rest components frozen.
The high-level encoder is detailed in Sec.~\ref{sec:UMAP}, while the low-level encoder comprises MLP backbone, diffusion prior~\cite{ramesh2022hierarchical}, and Versatile diffusion~\cite{Xu2023}.
We use a training dataset with image-text pairs.
Inspired by MindEye~\cite{Scotti2024}, the loss function for low-level $L_{image}$ and high-level $L_{text}$ embeddings are designed with integration of bidirectional CLIP~\cite{Radford2021clip} contrastive learning and MixUp~\cite{Zhang2017mixup} augmentation to improve generalization in the beginning of training stage, as described in the follows:
\begin{equation}
\resizebox{\dimexpr\linewidth-30pt\relax}{!}{$
\begin{aligned}
\mathcal{L}_{mix}= & - \sum_{i=1}^N [\lambda_i \cdot \log(\frac{\exp(\frac{p_i^* \cdot t_m}{\tau})}{\sum\limits_{m=1}^N\exp(\frac{p_i^* \cdot t_i}{\tau})})\\
&+(1-\lambda_i)\cdot \log(\frac{\exp(\frac{p_i^* \cdot t_{k_i}}{\tau})}{\sum\limits_{m=1}^N\exp(\frac{p_i^* \cdot t_m}{\tau})})] \\
& - \sum_{j=1}^N [\lambda_j \cdot \log(\frac{\exp(\frac{p_j^* \cdot t_j}{\tau})}{\sum\limits_{j=1}^N\exp(\frac{p_m^* \cdot t_j}{\tau})})\\
&+\sum_{\{l|k_l=j\}}(1-\lambda_l)\cdot \log(\frac{\exp(\frac{p_l^* \cdot t_{j}}{\tau})}{\sum\limits_{m=1}^N\exp(\frac{p_m^* \cdot t_j}{\tau})})],
\end{aligned}
\label{eq:Bimixco}$}
\end{equation}
where $\tau$ represents the temperature hyperparameter, $\lambda$ denotes the voxel-mixing coefficient obtained from a Beta distribution where $\alpha = \beta = 0.15$, and $N$ is the batch size.
The CLIP embedding $t_i$  of a ground-truth image or text, denoted as $y_i$, is obtained as:
\begin{equation}
t_{i}=\rm{CLIP}_{image/text}(y_i).
\end{equation}
The $p_i^*$ in Eq. \eqref{eq:Bimixco} represents a fMRI embedding derived from either a low-level encoder $f_{L}$ or a high-level encoder $f_{H}$, denoted as $f_{L/H}$, using the initial fMRI $X_i$. This process is obtained with MixUp augmentation~\cite{Zhang2017mixup} as the subsequent equations:
\begin{equation}
X_{\rm{mix}_{i,k_i}} = \lambda_i \cdot X_i + (1-\lambda_i) \cdot X_{k_i},
\end{equation}
\begin{equation}
p_i^* = f_{L/H}(X_{\rm{mix}_{i,k_i}}),
\end{equation}
After one third of the training stage, the BiMixCo loss for both $L_{image}$ and $L_{text}$ is replaced with soft contrastive loss to facilitate softmax probability distribution for improved knowledge distillation~\cite{Hinton2015}, expressed as:
\begin{equation}
\resizebox{\dimexpr\linewidth-30pt\relax}{!}{$
\mathcal{L}_{soft}=- \sum\limits_{i=1}^N \sum\limits_{j=1}^N[\frac{\exp(\frac{t_i \cdot t_j}{\tau})}{\sum\limits_{m=1}^N\exp(\frac{t_i \cdot t_m}{\tau})} \cdot  \log( \frac{\exp(\frac{p_i \cdot t_j}{\tau})}{\sum\limits_{m=1}^N\exp(\frac{p_i \cdot t_m}{\tau})} )].
\label{eq:softclip}$}
\end{equation}
where the $p_i$ represents fMRI embeddings derived from either low-level or high-level encoder $f_{L/H}$ through the following equation:
\begin{equation}
p_{i} = f_{L/H}(X_i).
\end{equation}
During {\bf inference}, we exclusively use pre-trained weights for fMRI extraction $p_{i} = f_{L/H}(X_i)$ and the subsequent 3D generation, relying entirely on fMRI data. 
This approach ensures that the model's inference is based purely on its understanding of fMRI data, without requiring any information such as the viewed image, nor external texts.

\section{Experiments}
\label{sec:results}

\subsection{Dataset}
We conduct extensive experiments on two publicly available fMRI datasets: the Natural Scenes Dataset (NSD)~\cite{allen2022massive} and the Generic Object Decoding (GOD)~\cite{horikawa2017generic} dataset.
NSD~\cite{allen2022massive} is comprised of high-resolution whole-brain fMRI measurements (1.8-mm isotropic resolution, 1.6-s sampling rate) from eight healthy adult subjects. These subjects were observed while viewing thousands of 2D colored natural images sampled from COCO dataset~\cite{lin2014microsoft}, spread across 30 to 40 scanning sessions. 
The two-stage training set includes 8,859 images and 24,980 fMRI trials, whereas the test set contains 982 images and 2,770 fMRI trials. 
To enhance data quality, we compute average fMRI responses for images that are presented more than once. 
GOD~\cite{horikawa2017generic} consists of fMRI recordings from five subjects viewing ImageNet stimuli, resulting in a training set of 1,200 images (8 images per 150 categories) and a test set of 50 images (1 from each of 50 categories). Each subject’s data amounts to 1,200 training and 1,750 testing fMRI trials.
While offering no comprehensive annotations for ROI-based analysis as NSD, 
this dataset enables generalization and robustness evaluation of our method across different stimuli and scanning conditions.

The fMRI signals of NSD are preprocessed and split into two specialized regions using predefined masks: the NSD General Region-of-Interest (ROI) mask and the Medial Temporal Lobe (MTL) mask, both at a resolution of 1.8 mm. The NSD General ROI mask encompasses regions such as V1, V2, V3, and V4, presenting a spectrum from the early visual cortex to higher visual areas, with an emphasis on low-level information. In contrast, the MTL mask encompasses brain regions including the hippocampus, amygdala, and parahippocampal regions, which are more attuned to high-level information.

\begin{figure*}[!t]
    \centering
\includegraphics[width=\linewidth]{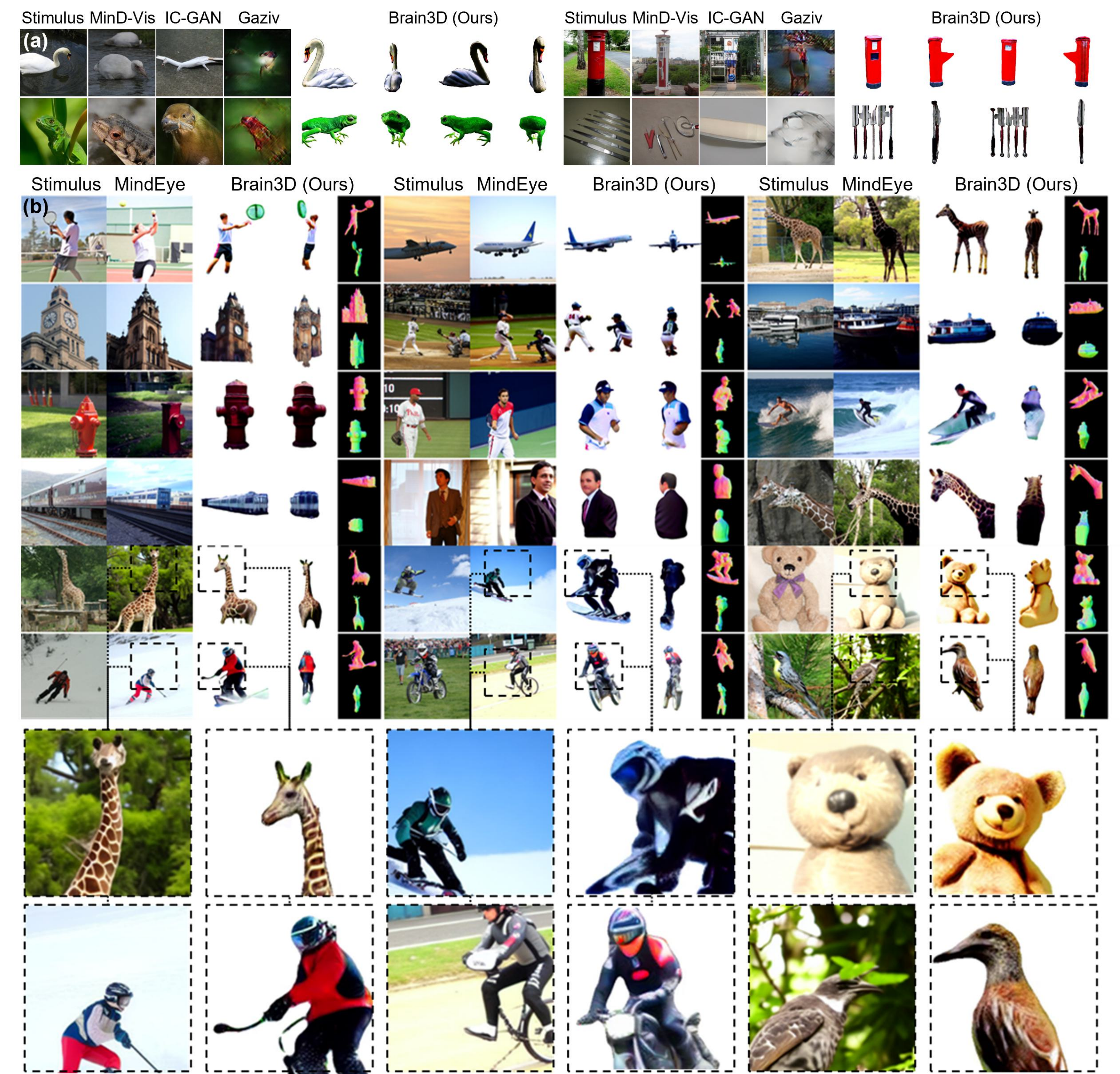}
    \caption{
    {\bf \ourmodel{} generates finer 3D visuals from fMRI under 2D stimuli compared with previous methods.}
    (a) Evaluation on the Generic Object Decoding (GOD)~\cite{horikawa2017generic} dataset. The first column is the 2D stimuli displayed to the subjects. The second, third, and fourth columns exhibit the 2D reconstruction from MinD-Vis~\cite{chen2023seeing}, IC-GAN~\cite{ozcelik2022reconstruction}, and Gaziv~\cite{gaziv2022self}. The last columns display the multi-view visualization of our generated 3D visuals from fMRI. Our method not only achieves the best quality of fMRI-based generation, but also uniquely infers 3D geometry.
    (b) Evaluation on the Natural Scenes Dataset (NSD)~\cite{allen2022massive}.
    For each case, the first column displays the 2D stimuli presented to the subjects. The second column presents 2D reconstruction from MindEye~\cite{Scotti2024}. The three and fourth column exhibits two distinct views of our 3D visuals generated from fMRI, followed by the final column depicting the surface normals of our generation, yielding higher quality output than MindEye~\cite{Scotti2024}. Zoomed-in visualizations in the last two rows demonstrate our better quality and more accurate fMRI-based generation.}
    \label{fig:3}
\end{figure*}

\subsection{Evaluation metrics}
\label{sec:evaluation}
To analyze the generation quality, we introduce four fMRI-to-3D evaluation metrics: 3D Inception~\cite{Szegedy2016Inception}, 3D CLIP~\cite{Radford2021}, 3D EfficientNet~\cite{tan2023Efficientnet}, and 3D Swav~\cite{caron2023SwAV}. 
These metrics are designed to overcome the limitations of traditional pixel-by-pixel image reconstruction metrics like PixelCorrelation~\cite{Potetz2003}, which lacks robustness against rotations and translations in 3D space, and are built upon well-established metrics in previous neural decoding researches~\cite{Chen2023, Scotti2024} to ensure reliability.
For comprehensive evaluation, we construct a level-2 icosahedron with a 2.2 radius around the generated 3D object and capture the 3D object from all the icosahedron vertices to achieve multi-view coverage, similar to the approach in the text-to-3D benchmark~\cite{he2022T3bench}.
The metrics 3D Inception and 3D CLIP, utilizing the final pooling layer of InceptionV3~\cite{Szegedy2016Inception} and CLIP model~\cite{Radford2021} respectively, assess the accuracy of two-way identification between original image embeddings and corresponding brain embeddings. Consequently, larger 3D Inception and 3D CLIP represent higher matching accuracy in features between ground truth images and generated 3D objects, indicating a well extracted features from fMRI.

On the other hand, 3D EfficientNet and 3D Swav employ EfficientNet-B1~\cite{tan2023Efficientnet} and SwAV-ResNet50~\cite{caron2023SwAV} to calculate the average correlation distance between the extracted features. Lower correlation distance represents higher correlation between ground truth images and generated 3D objects, representing better reconstruction. 

In order to address view inconsistency in the final quality evaluation, we implement a regional convolution mechanism to smooth the scores across the local region, which treats the icosahedron as a graph composed of vertices and edges and performs mean pooling on this graph, employing a recursive formula:
\begin{equation}
s_i^{(t+1)}=\frac{1}{\left| N(i) \right|+1} ( s_i^{(t)}+\sum_{j\in N(i)} s_j^{(t)}).
\label{eq:metric}
\end{equation}
The highest score among all viewpoints is then selected as the ultimate quality score for the 3D generation.

\begin{figure*}[!t]
    \centering
\includegraphics[width=\linewidth]{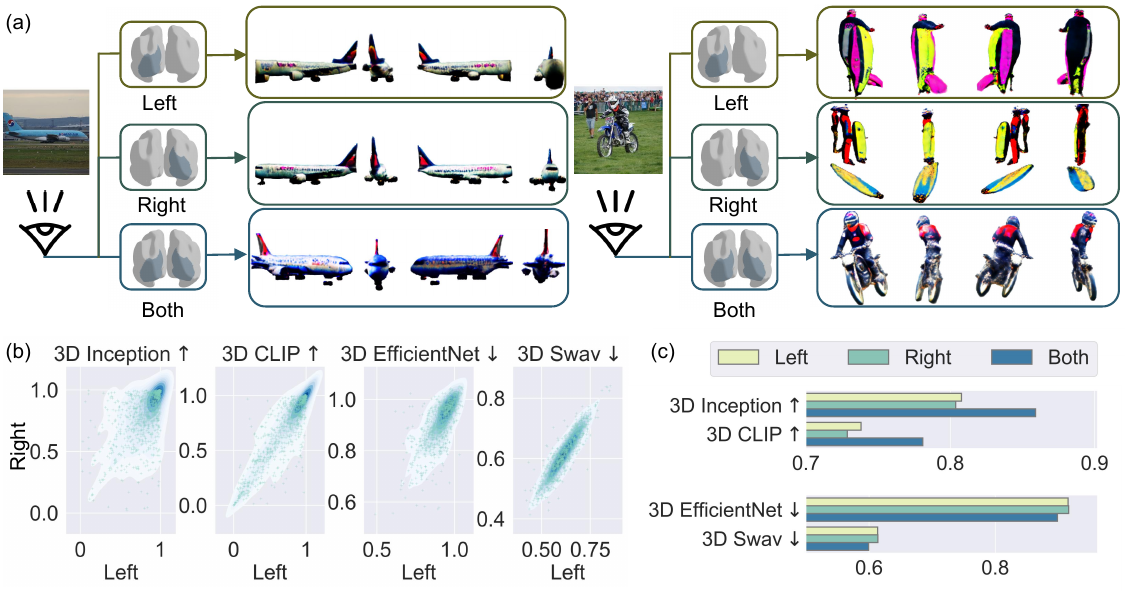}
    \caption{
    \textbf{Collaborative property of the left and right hemispheres with \ourmodel.} (a) 
    To examine the difference and interaction between the human brain's two hemispheres, 
    we conduct specialized experiments assessing 3D generation quality using fMRI from either hemisphere or both.
    (b) 
    The function of the left and right hemispheres varies significantly for different specific objects under various metrics. In general, the left favors finer details and intricate structures, while the right focuses more on overall shape and silhouette.
    (c)  
    Notably, using a combination of data from both hemispheres led to an enhanced generation. This synergistic effect was evident in improved performance across various metrics.}
    \label{fig:4}
\end{figure*}

\begin{figure*}[htbp]
    \centering
\includegraphics[width=0.94\linewidth]{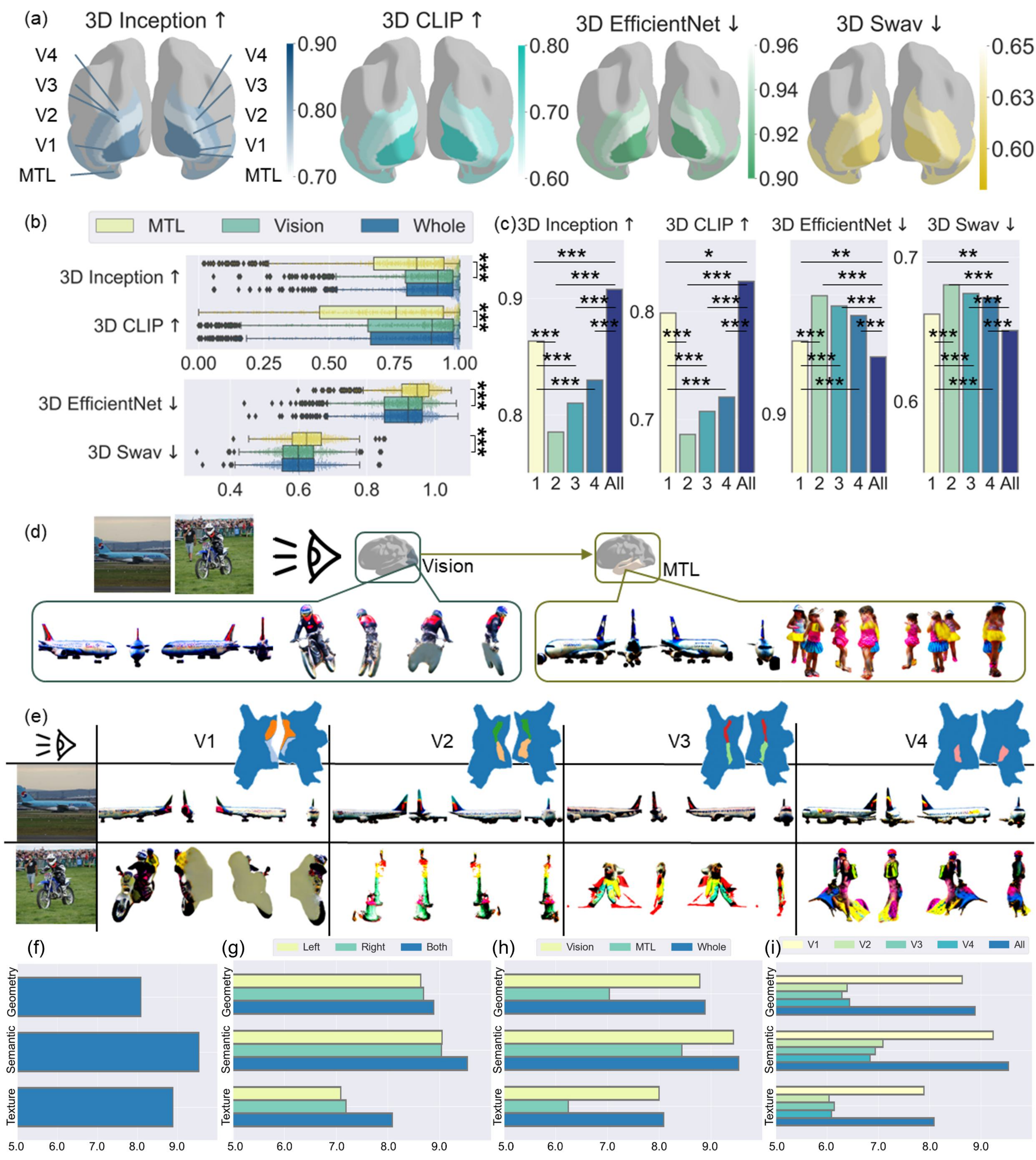}
\caption{\textbf{Visual region predominates in comprehension.} 
    (a) 
    Overall, visual regions outperform the medial temporal lobe (MTL) regions under varying metrics.
    Further, V1 contributes the most among visual regions.
    (b) The vision region demonstrates a higher 3D generation quality with elevated performance. 
     The significance level for 
MTL vs vision is: 3D Inception:$ p=1e^{-23}$, 3D CLIP: $p=1e^{-16}$, 3D EfficientNet $p=1e^{-22}$, 3D Swav $p=3e^{-15}$.
    Performance of each generated object in these criteria is represented by crosses (+). 
     (c) V1 Regions exhibit better performance than V2, V3 and V4 in generation.
    Significance levels are indicated as $*: p<1e^{-2}, **: p<1e^{-3}, ***: p<1e^{-4}$. 
    (d) MTL region tends to extract semantic visual concepts, while the visual cortex concentrates on textures and silhouettes.  
    (e) V1 dominates in inferring the object silhouette, whilst V2, V3, and V4 are more focused on detailed aspects of texture.
    (f) Our Brain3D exhibits a high level of semantic consistency along with considerable texture and geometric consistency. (g) Left and right hemispheres exhibit similar consistency scores.  (h) The visual region dominates over MTL. (i) The V1 region outpaces the others clearly.
    }
    \label{fig:5}
\end{figure*}

\begin{figure*}[!t]
    \centering
\includegraphics[width=\linewidth]{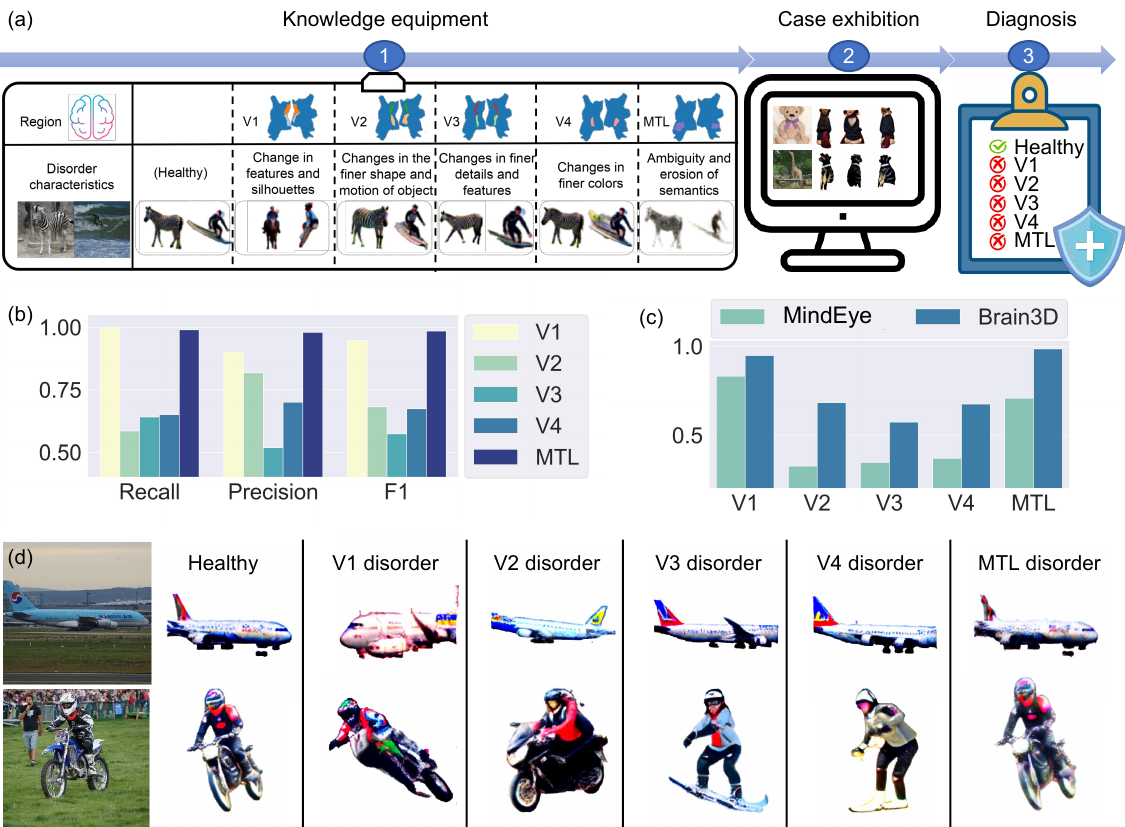}
    \caption{\textbf{Illustration of \ourmodel{} to process fMRI for clinical diagnosis.} 
    (a) In non-expert diagnosis, participants are first equipped with knowledge about the function with varying visual system regions (V1, V2, V3, V4, MTL) and a number of case study for their diagnosis practice. 
    We show cases both with and without disorders for information.
    (b) Regional disorder diagnosis performance by non-professionals using our Brain3D (chance level: 0.17).
    (c) 3D representation of \ourmodel{} achieves improved F1 diagnosis score than 2D model.
    (d) Visual example of disorder diagnosis with \ourmodel{}.}
    \label{fig:6}
\end{figure*}

\begin{figure*}[!t]
    \centering
\includegraphics[width=\linewidth]{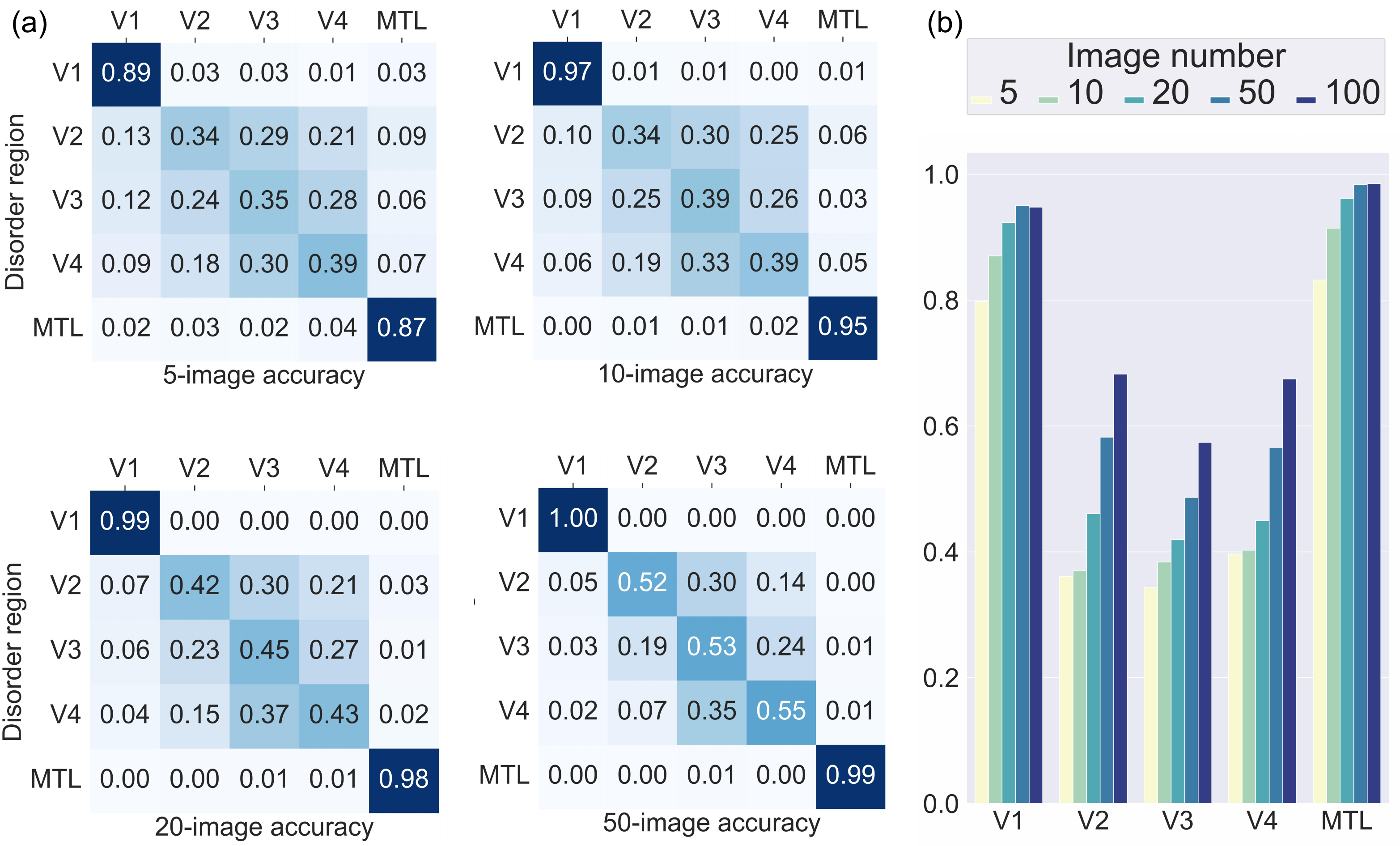}
    \caption{\textbf{Advancing efficient diagnosis with limited images.} 
    (a) Our study further conducted experiments where participants were given restricted sets of visualizations to determine the feasibility of low-cost diagnosis. This approach underscores the remarkable diagnostic proficiency within the V1 and medial temporal lobe (MTL) regions, even when using a limited quantity of images. We present heatmaps that detail diagnostic accuracy, utilizing varying sets of images - namely, batches of 5, 10, 20, and 50 images per patient. The results demonstrate that participants could accurately discern disorders in the V1 and MTL regions, despite the limited number of images available. (b) Bar-plot presents the recall of V1, V2, V3, V4, and MTL region diagnosis under different image number. It suggests that reliable diagnostic conclusions of V1 and MTL can be drawn with fewer images, leading to a substantial reduction in time and resources required for diagnosis. This efficiency is especially critical in clinical settings, where swift and accurate diagnosis is paramount. }
    \label{clinic_sample}
\end{figure*}

\subsection{Qualitative evaluation} 
To assess the generalizability and practical utility in clinical settings of \ourmodel{}, we perform comprehensive comparisons with previous 2D methods~\cite{Scotti2024, chen2023seeing, ozcelik2022reconstruction, gaziv2022self}.
We could not perform quantitative comparison as all previous 2D methods 
do not output 3D generation as ours.
As shown in Fig.~\ref{fig:3},
from the fMRI data \ourmodel{} extracts rich appearance, semantic, and geometric information, presenting a high degree of consistency with the stimuli images.
In the left columns of Fig.~\ref{fig:3}a, the generated swan and lizard from \ourmodel{} not only show the highest semantic and texture consistency with the stimulus, but also preserve 3D geometry details inferred from brain signals.
Despite limited observation with the giraffe, skier and teddy bear in the three cases (see the last two rows of Fig.~\ref{fig:3}b), our bio-inspired model can process 3D details and comprehend their semantics beyond what is immediately visible in the 2D image, achieving finer semantic details than  MindEye~\cite{Scotti2024}. As shown in Fig.~\ref{fig:4}c with both hemispheres, our model also achieves a significant quantitative neural decoding performance across all samples in the testing set.
This result underscores the remarkable human's neural processing capability in constructing intricate 3D structures by interpreting and internalizing 3D geometric information from 2D visual stimuli~\cite{Georgieva2009,Groen2019}.

\subsection{Brain region functionalities}

\noindent\textbf{Enhanced visual quality through collaboration between the left and right hemispheres.} 
We investigate whether our model presents biological consistency in terms of regional functionalities.
We first examine the roles of the left and right brain hemispheres, as well as their collaboration during the 3D object reconstruction process (Fig.~\ref{fig:4}a). 

Specifically, we observe that the two hemispheres make different holistic contributions to object perception, due to exhibiting distinct performances for specific objects (Fig.~\ref{fig:4}b). The left hemisphere tends to depict finer details and intricate structures in object visualization, while the right hemisphere better captures the overall shape and silhouette. These results align with previous discoveries that the left hemisphere is involved in detail-oriented tasks while the right hemisphere in more holistic tasks~\cite{Iaccino2014Left}. 
However, our results challenge the notion of strict functional division. We found notable overlaps and interactions between the two hemispheres. Remarkably, numerous objects rendered by the right hemisphere, as illustrated in Fig.~\ref{fig:4}a, also exhibit intricate textures—typically a characteristic attributed to the left hemisphere.  Conversely, the left hemisphere demonstrated capability in detecting overarching shapes, which is usually a strength of the right hemisphere. 

Despite these different functional tendencies, our quantitative analysis, using metrics like 3D Inception, 3D CLIP, 3D EfficientNet, and 3D Swav in Fig.~\ref{fig:4}c, shows that both hemispheres contribute similarly to 3D object perception.
Their collaboration significantly contributes to the quality of 3D objects (Fig.~\ref{fig:4}c).
This result echos the biological discovery on the interaction between the left and right brain~\cite{Corballis2014Left,Iaccino2014Left}.

\noindent\textbf{Dominance of V1 visual regions in feature and silhouette processing.}
We further investigate the functionalities of the various visual regions and the medial temporal lobe (MTL) region (Fig.~\ref{fig:5}a).

Previous studies have demonstrated that visual cortex~\cite{Milner2006visual} is predominantly involved in processing color and texture information, whereas the medial temporal lobe (MTL) region~\cite{Squire2004MTL,  Eichenbaum2007MTL} is more focused on semantic information. 
In comparison, our research shows that the visual cortex exhibits higher visual decoding ability than MTL region (Fig.~\ref{fig:5}b). The significance level refers to the probability of error when rejecting the null hypothesis that visual cortex shares the similar performance as MTL region. Specifically, the visual cortex captures local fine-grained visual information such as the ``windows''  of an airplane and the ``motorcycle rider'', 
while the MTL region captures holistic shapes like ``airplane skeleton'' and ``person'' (Fig.~\ref{fig:5}d). This visualization  
highlights higher-level 3D-shape-related concepts while foregoing pixel-level details. 
Our extensive analysis illustrated in Fig.~\ref{fig:5}f shows a high level of semantic consistency along with considerable texture and geometric consistency.
This discrepancy may be attributed to the inherent nature of human visual processing, where the brain may not emphasize all texture details, particularly in background or less prominent areas.
In regional analysis, we observe that hemispheric symmetry is evident as the left and right hemispheres exhibit similar scores across texture, semantic, and geometric metrics (Fig.~\ref{fig:5}g), indicating balanced processing. Furthermore, significant visual cortex dominance is apparent in perceptual processes (Fig.~\ref{fig:5}h), demonstrating a superior capability to integrate geometry, semantics, and textures compared to the MTL, which, while showing reasonable semantic consistency, lacks consistency in geometry and texture. 
This finding offers a novel perspective on how the MTL may contribute to the formation and recall of long-term memories, highlighting its role in abstracting complex visual information at the semantic level.

In line with earlier researches~\cite{Tong2003primary}, V1 region exhibits predominant ability in visual representation (Fig.~\ref{fig:5}c) among various visual regions. 
To further break down, V1 region plays a pivotal role in handling elementary visual information, including edges, features, and color.
The V1 region's essential function in managing elementary visual information is underscored by its decoding results (Fig.~\ref{fig:5}i), which reveal more substantial semantic, geometric, and texture consistency over other regions.
V2 and V3 regions play a role in synthesizing various features into complex visual forms,
while V4 region focuses on basic color process at coarse level (Fig.~\ref{fig:5}e).
However, our research brings to light a nuanced understanding of the interdependence within the visual cortex. Specifically, we demonstrate the reliance of the V2, V3, and V4 regions on the V1 region for constructing 3D vision. This is evident from the disrupted 3D geometries observed in  Fig.~\ref{fig:5}e. 
Moreover, scenes decoded from the V2, V3, and V4 regions exhibit only moderate consistency in texture, semantics, and geometry (Fig.\ref{fig:5}g). 
This cross-region reliance 
likely stems from the fact that these regions process information initially received by V1~\cite{Hubel1979brain}, indicating a foundational dependence on V1 for effective visual processing.
These findings emphasize the collaborative and interactive roles of these brain regions in forming a comprehensive visual representation.
This result is thus biologically consistent to great extent.

\begin{figure*}[!ht]
    \centering 
    \setlength{\tabcolsep}{0mm}{
    \begin{tabular}{c} 
    \includegraphics[width=0.98\textwidth]{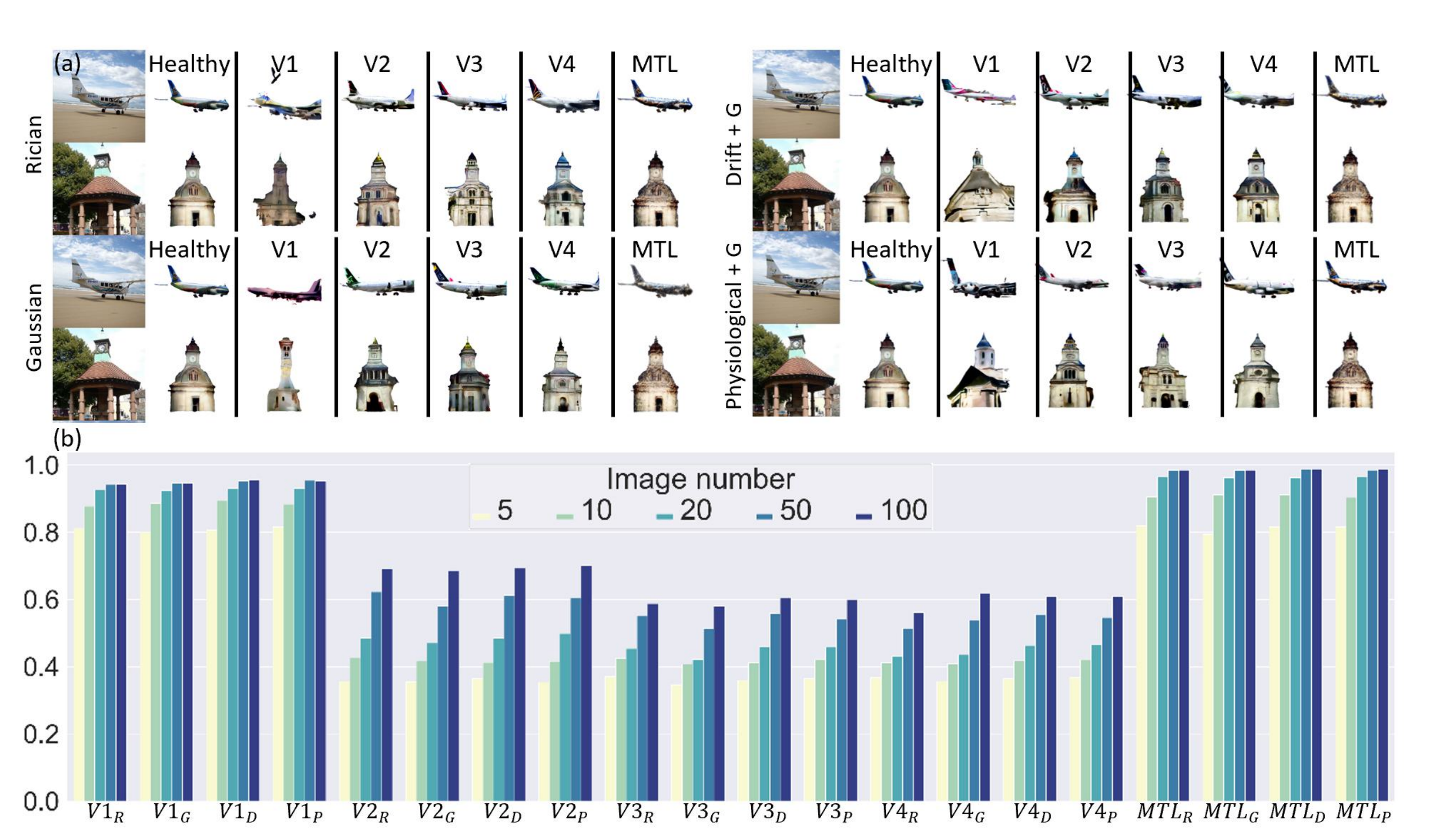} \\
    \end{tabular}
    } 
\caption{\textbf{blation study on simulating regional disorders}. Here, we evaluate the impact of different noise models, including Rician (R), Gaussian (G), Drift (D), and Physiological (P), on our simulations of regional disorders. We show that (a) V1 is the most sensitive region to the specific noise model used, while V2, V3, and V4 exhibit detailed variations in shape, local features, and color, and the MTL region consistently displays similar blurring effects. (b) When assessing the diagnostic recall performance for all regions across a range of 5 to 100 diagnostic images, our method demonstrates high consistency and stability across all noise models.}
\label{fig:ablation_noise}
\end{figure*}

\subsection{Clinical evaluation}
\label{sec:method_simulation}
Our model has the ability to translate complex fMRI signals into easily understandable multi-view object images. This empowers individuals, regardless of their professional background, to undertake disorder diagnosis with minimal training through a straightforward tutorial. This substantially lowers the barrier to entry for this task.

To simulate the disorders of varying brain regions,
we adopt the general fMRI simulation techniques~\cite{Welvaert2014} that introduce Gaussian noises.
We randomly select samples from the NSD test set for disorder simulation.
For each sample, we apply independent Gaussian noises with a variance of 3.0 over five brain regions: V1, V2, V3, V4, and MTL, respectively.

In our diagnosis tutorial we show a participant several examples including the original image presented to a patient, and our model's output under the healthy situation and regional disorder in V1, V2, V3, V4, and MTL, along with a textural description of each disorder's characteristics.
This tutorial can be easily conducted before the user study.

In our experiment, we invited 100 volunteer participants, 
each educated simply by the above tutorial. 
None of them possessed prior expertise in neuroscience or brain disorder diagnosis with fMRI. 
The diagnosis task is to detect which brain region has disorders by
comparing the original image and the generated images by our model.
For diagnosing a specific patient, we use a various number $\{1, 5, 10, 20, 50, 100\}$ of images and fMRI pairs associated.
We evaluate the diagnosis performance by 
the metrics of precision, recall, and F1 score.

\noindent\textbf{Visual region clinical diagnosis}
Motivated by the high biological consistency in terms of regional functions, we examine the usefulness of \ourmodel{} in diagnosing disordered brain regions by designing a series of simulation based experiments (Fig.~\ref{fig:6}a). To compare performance of \ourmodel{} and 2D models, we designed three branches of experiments: 2D model MindEye~\cite{Scotti2024}, \ourmodel{} front view and  \ourmodel{} back view. In each branch, the participants were tasked with determining the impacted brain region of the patient from V1, V2, V3, V4, and MTL by observing 3D visuals generated from their fMRI signals.
Their average correctness of judgement is represented as evaluation of clinical diagnosis.
Notably, almost perfect diagnosis is achieved for the disorders with the V1 and MTL regions (Fig.~\ref{fig:6}b). However, distinguishing the disorders among V2, V3, and V4 regions presents more challenges, as reflected by the incidence of incorrect diagnoses in these areas.  This difficulty is attributed to the significant overlap and interaction in their processing roles, as they all engage in visual information processing following the V1 region.
Nonetheless, when a disorder was localized to a single region among V2, V3, or V4, participants generally succeeded in identifying that the issue stemmed from one of these three regions, despite the nuanced differences in their functionalities. 
Diagnosis among all regions in \ourmodel{} outperforms previous 2D methods (Fig.~\ref{fig:6}c). 
We provide visual examples for region disorder diagnosis (Fig.~\ref{fig:6}d).
V1 disorder tends to cause significant distortion to the output, consistent with the earlier finding about its critical role in processing basic visual information.
Disorder in V2 region leads to changes in the detailed shape and motion of objects, e.g., orientation of the airplane and shape of the motorcycle.
V3 region primarily affects the details and movements, e.g., the change from motorcycle to rider. Disorder in V4 region might cause color shift and variation.
Instead, MTL's disorder could be concerned with high-level semantics, e.g., increased ambiguity in object appearance.
See Sec.~\ref{sec:method_simulation} for more details.

\noindent\textbf{Cost-effective diagnosis}
To explore more cost-effective diagnostic methods, our study extended to experiments where participants were provided with a limited number of visualizations for brain disorder diagnosis. This approach was driven by the desire to ascertain if accurate diagnostics could be attained with fewer resources such as time and cost. 
The results were significant, as demonstrated in Fig.~\ref{clinic_sample}: even with a reduced number of images, participants were able to accurately tell disorders in the V1 and MTL regions,
significantly cutting down the time and resources usually needed for thorough diagnoses. 
Specifically, even with as few as 10 images (Fig.~\ref{clinic_sample}a), participants achieved over 95\% of diagnostic accuracy for disorders in the V1 and MTL regions. This implies that by utilizing merely 10\% of the potential images from patients, we can still reach a plausible diagnisis of V1 and MTL region thereby saving almost 90\% of the time and effort as required in our full diagnostic process.

\noindent\textbf{Diagnosis on complex simulations} 
To explore diagnosis through more complex neurological conditions, we further extend our experiments with Rician~\cite{wegmann2017bayesian, gudbjartsson1995rician} (with a variance of 3.0), Drift~\cite{gupta2022neural} (with a drift rate of 0.1), and Physiological~\cite{biswal1996reduction,madl2022mri} (with a fluctuation intensity of 0.5) simulations across all five brain regions (V1, V2, V3, V4, and MTL). Both Drift and Physiological simulations are further combined with Gaussian noise to mimic more complex neurological conditions~\cite{gupta2022neural,biswal1996reduction,madl2022mri}.
Figure~\ref{fig:ablation_noise}(a) shows how different simulated disorders impact brain regions. For V1, each simulation type creates distinct distortions: Rician makes results vague, Gaussian distorts shapes, Drift breaks up textures, and Physiological adds irrelevant features. Despite these variations, all V1 simulations severely disrupt basic visual information.
In V2, all simulations lead to changes in detailed shapes. For V3, they slightly alter fine details. For V4, disorders primarily change detailed colors, like making the airplane whiter and the clock tower bluer. Importantly, even with these changes, the overall outlines of objects like airplanes and clock towers remain largely visible across V2, V3, and V4, highlighting these regions' supportive roles.
Disorders in the MTL region, across all simulations, increase the ambiguity of object appearance, showing its link to high-level meaning. Figure~\ref{fig:ablation_noise}(b) confirms that our diagnostic method provides reliable conclusions across all simulations.

\begin{figure*}[!ht]
    \centering
\includegraphics[width=\linewidth]{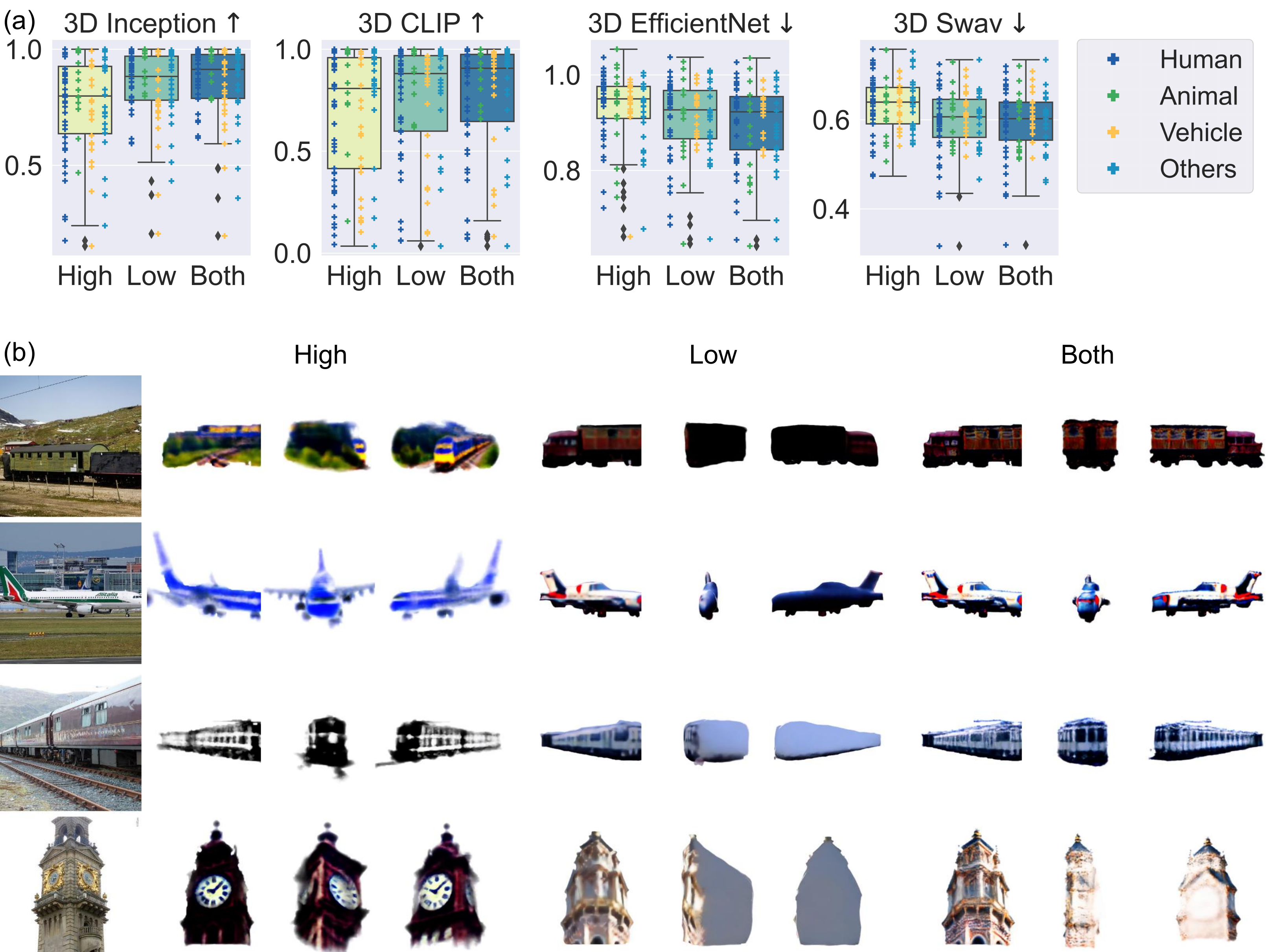}\caption{\textbf{Both low-level and high-level embeddings can be detrimental in visual processing when used independently.}
    (a) Integration of both Low-Level and High-Level is beneficial for human brains to better perceive objects. Boxplots exhibit the median, the 25th, and 75th percentiles as box edges among the four metrics.  Crosses (+) represent the scores of each image with the category displayed to participants.
    (b) High-level embedding in fMRI concentrates on abstract semantic information, while low-level embedding in fMRI focuses on color and texture information. The synergy of these two embeddings results in a more holistic perception of objects in the human brain, encapsulating both semantic depth and textural detail. 
    }
    \label{high_low}
\end{figure*}

\begin{figure*}[!t]
    \centering
\includegraphics[width=\linewidth]{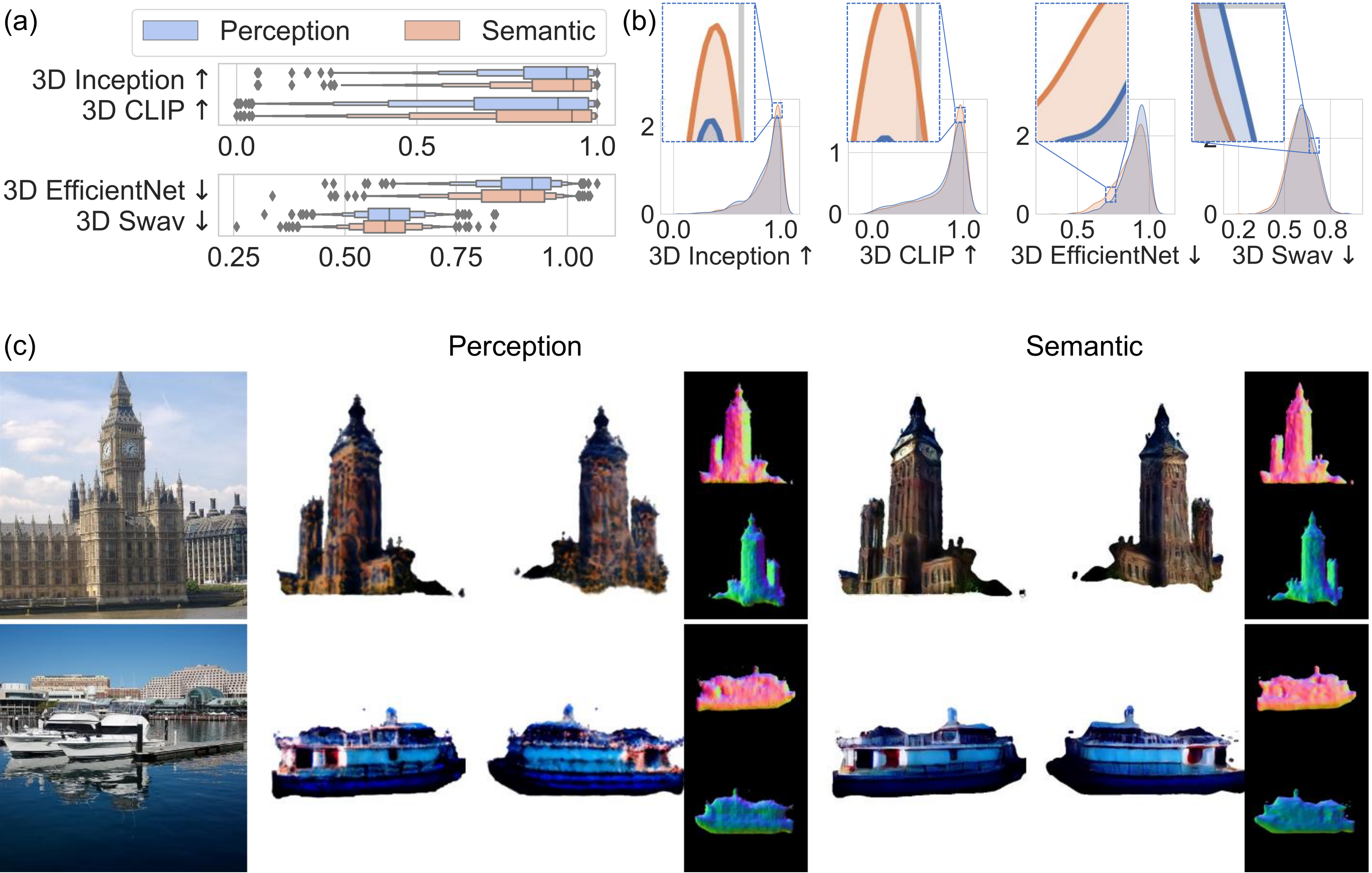}
    \caption{\textbf{ Enhanced generation quality through semantic stage.} (a) The semantic stage outperforms the perception stage in all evaluation metrics in every quantile. Boxen plot shows 5\%, 15\%, 25\%, 50\%, 75\%, 85\%, and 95\% percentile. (b) During the semantic stage, there is a noticeable increase in the proportion of high 3D Inception and 3D CLIP scores, along with a leftward shift in the 3D EfficientNet and 3D Swav scores. This shift indicates a more refined and accurate generation of 3D objects.  (c) 3D objects generated in the semantic stage exhibit a noticeably higher fidelity compared to those from the perception stage, showcasing the enhanced capabilities of the semantic processing.}
    \label{ablation}
\end{figure*}

\begin{figure*}[htbp]
    \centering
\includegraphics[width=\linewidth]{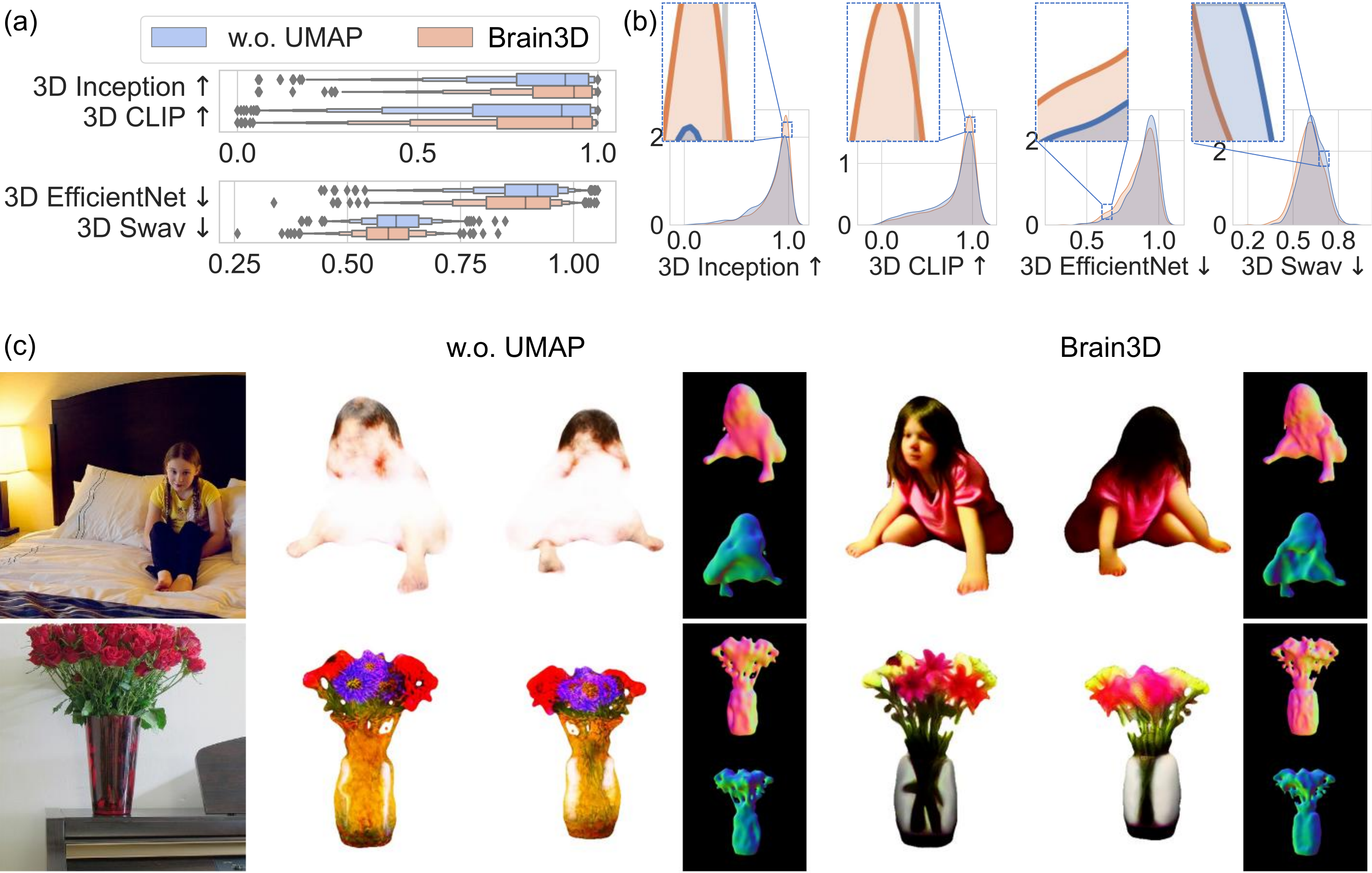}
    \caption{\textbf{ Enhanced generation quality through UMAP projection.} (a) UMAP projection improves generation quality among all metrics in every quantile.  (b) Utilizing UMAP projection not only achieves higher 3D Inception and 3D CLIP scores but also contributes to lower 3D EfficientNet and 3D Swav scores. This outcome demonstrates the projection's ability to refine and improve the overall quality. (c) The employment of UMAP projection significantly enhances textural stability in fMRI-generated content, pointing to its efficacy in producing more consistent and reliable outputs.  (d) Ablation on the number of components of UMAP. The setting of $N_{neighbor}=50$ (our default) achieves the best recovery of 3D geometry and textures. An extremely sparse setting of $N_{neighbor}=1$ results in disrupted semantic representation, while increasing $N_{neighbor}$ to $200$ and $1000$ leads to no improvement but slightly more vague texture.} 
    \label{ablation2}
\end{figure*}

\begin{figure*}[!ht]
    \centering
\includegraphics[width=\linewidth]{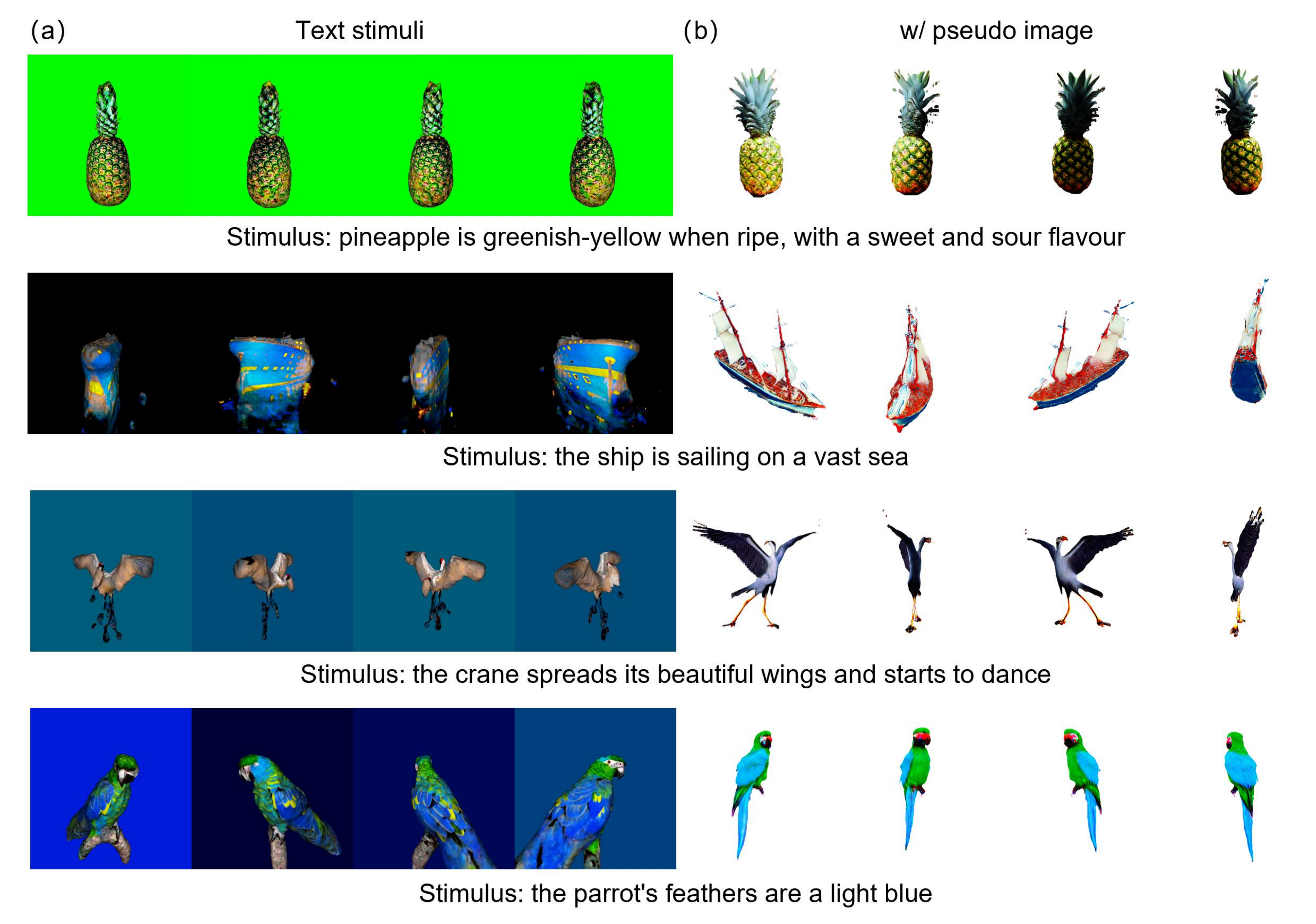}\caption{ 
\textbf{Evaluation on brain signals derived from text-based imagination and in the complete absence
of image stimuli.} 
(a) With only text stimuli based on fMRI, our model still successfully generates visuals that exhibit
a commendable alignment with the textual prompts.
(b) After incorporating pseudo images from an off-the-shelf text-to-image generator, our model achieves an enhanced translation from fMRI signals to more detailed and refined visual representations.
    }
    \label{fig:ablation_more}
\end{figure*}

\subsection{Ablation analysis}
\noindent\textbf{Ablation on guidance}
Our perceptual pipeline employs Score Distillation Sampling (SDS)~\cite{poole2022dreamfusion} with the inclusion of both low-level and high-level guidance. The integration of both low-level and high-level guidance achieves the best performance across all categories (Fig. \ref{high_low}a).
As shown in Fig. \ref{high_low}b, when solely relying on high-level guidance, the representations of the train, airplane, and clock tower appear significantly blurred, displaying only coarse shapes.
These coarse shapes stem from the geometric and semantic information embedded within the high-level representation.
Low-level guidance primarily captures the color and texture attributes of the objects. Consequently, the outcomes with low-level guidance (see the first column), featuring the front view, align well with the provided image, while other views fall short in terms of shape and semantic fidelity.
The combination of both high-level and low-level guidance results in significantly improved representations of the airplane, train, and clock tower, where both the semantic and textural structures closely resemble those of the testing image. 
\noindent\textbf{Ablation on the Semantic Stage}
The Semantic Stage within our framework is designed to extract finer details from fMRI embeddings, enhancing its contribution to 3D generation. 
After the Semantic Stage the detailed textures of the building in the first row and the outline of the steamship in the second row in Fig.~\ref{ablation}c exhibit much greater clarity. These details from fMRI are challenging to represent in a coarse NeRF but can be effectively distilled within a 3D mesh. 

\noindent\textbf{Ablation on UMAP projection}
The left four-column section in Fig.~\ref{ablation2}c  displays the results obtained using high-level embeddings directly extracted from fMRI without the application of UMAP projection. In this scenario, the target distribution $p_{\phi}(\mathbf{z}_t |\mathbf{y})=\int p{\phi}(\mathbf{z}_t |c,\mathbf{y})p(c) dc$ is optimized with respect to $\mathbf{y}$. The multi-view integration $\int p{\phi}(\mathbf{z}_t |c,\mathbf{y})p(c) dc$ integrates $\mathbf{y}$ with high diversity and substantial noise, resulting in oversaturated and inconsistent generation outcomes.
Following the incorporation of our designed UMAP projection, illustrated in the right four-column section in Fig.~\ref{ablation2}c, the high-level information $\mathbf{y}$ undergoes the transformation into a series of stable and consistent embeddings $\mathbf{y}_i$. Consequently, the target distribution is updated as
$p_{\phi}(\mathbf{z}_t |\mathbf{y}_i)=\int p{\phi}(\mathbf{z}_t |c,\mathbf{y}_i)p(c) dc$.
This adjustment steers the 3D optimization process toward a more stable and consistent direction in each iteration. Despite setting different guidance $\mathbf{y}_i$ sampled from the projection weight in each denoising iteration, the stability of the denoising step effectively ensures the attainment of a final stable generation.
Fig.~\ref{ablation2}d further shows dedicated analysis for the parameter choices in our UMAP projection through a greedy search for the parameter $N_{neighbor}$ in the range of $[1, 1000]$. 
Our setting with $N_{neighbor}=50$ (the default) achieves optimal recovery of 3D geometry and textures. This selection balances the preservation of local versus global semantics, ensuring a detailed and coherent representation. The setting of $N_{neighbor}=1$ results in significant disruption of semantic representation, highlighting an over-emphasis on specific local semantic choice. Largely increasing $N_{neighbor}$ to $200$ and $1000$ raises computational costs significantly whilst results in vague textures and no marked improvement over the default setting. A plausible reason is that using too many neighbors may lead to dilution and conflicts arising from noisy semantic embeddings.

\noindent\textbf{Ablation on visual stimuli}
To explore the purely imaginative scenario of generating objects without relying on reference images, we leveraged the TMNRED dataset \cite{bai2025tmnred}. This dataset is particularly well-suited for this purpose as participants' brain signals were collected exclusively under text description stimuli, with no accompanying visual input. Our Brain3D model is designed to process generic brain signals (e.g., EEG), regardless of the format of the visual stimuli used during data collection.
As demonstrated in Figure~\ref{fig:ablation_more}(a), our model successfully generates visuals that exhibit a commendable alignment with the textual prompts, even when conditioned solely on brain signals derived from text-based imagination and in the complete absence of image stimuli. This capability underscores our model's unique ability to directly decode and reconstruct mental imagery from brain activity, highlighting its potential to tap into the human imagination guided by text, past experience, and knowledge.
Furthermore, we investigated the effect of incorporating a synthetic visual prior. When our model was additionally equipped with a paired training image generated by Stable Diffusion~\cite{rombach2022high} (conditioned on the original text), Figure~\ref{fig:ablation_more}(b) shows that finer object details can be recovered. This suggests that our model can effectively integrate such a generative visual prior to enhance the translation from fMRI signals to more detailed and refined visual representations.

\section{Conclusions}

Understanding how human vision system extracts and processes the physical world is a critical question in neuroscience.
For the first time, this research models the underlying mechanisms of human visual perception by mapping fMRI of human brain activities to the corresponding 3D objects.  
Our research highlights the human brain's remarkable capability in perceiving 3D-related high-level information from 2D visual stimuli. This perception is facilitated by the synergistic functions of both brain hemispheres and the intrinsic collaboration between different functional regions.
We have attempted to simulate these intricate mechanisms of human visual perception by innovating a machine learning approach with great potentials for real-world applications.
Extensive experiments validate the performance and excellence of our method
over the prior alternative.
By enabling deeper and more direct analysis of visual perception,
our study opens up a more advanced and convenient approach to the clinical diagnosis of brain disorders.

\section{Data Availability Statement}
The datasets generated and/or analysed for modeling during the current study are available in: Natural Scenes Dataset (NSD)~\cite{allen2022massive}  (\href{https://natural-scenes-dataset.s3.amazonaws.com/index.html}{https://natural-scenes-dataset.s3.amazonaws.com/index.html}). The dataset generated and analyzed in brain region analysis and diagnosis will be made available upon request after the peer-review process. Interested researchers can request access by contacting the corresponding author via the journal's communication channels.

\section*{Acknowledgments}
This work was supported in part by National Natural Science Foundation of China (Grant No. 62376060).

\bibliographystyle{spbasic}      
\bibliography{main}   


\end{document}